\pdfoutput=1

\documentclass[11pt]{article}

\usepackage[final]{acl}
\usepackage{diagbox}
\usepackage{color}
\usepackage{tcolorbox}
\usepackage{CJK}
\usepackage{adjustbox}
\tcbset{width=0.9\textwidth,boxrule=0pt,colback=red,arc=0pt,auto outer arc,left=0pt,right=0pt,boxsep=5pt}
\usepackage{times}
\usepackage{latexsym}

\usepackage[T1]{fontenc}

\usepackage[utf8]{inputenc}

\usepackage{enumitem}

\usepackage{microtype}
\usepackage{adjustbox}
\usepackage{pifont}
\usepackage{multirow}
\usepackage{url}            %
\usepackage{booktabs}       %
\usepackage{xcolor}         %
\usepackage{colortbl}
\usepackage{graphicx}
\usepackage{subcaption}
\usepackage{wrapfig}
\usepackage{bbding}
\usepackage{ulem}
\usepackage{amsmath} 
\usepackage{longtable}
\usepackage{tabularx} 
\usepackage{lipsum}

\usepackage{inconsolata}

\newcommand{\done}[1]{{\textcolor{blue}{[DONE]}}}

\newcommand{\modified}[1]{{\textcolor{Maroon}{[MODIFIED]}}}
\newcommand{\modifying}[1]{{\textcolor{Maroon}{[MODIFYING]}}}

\title{CoGenesis: A Framework Collaborating Large and Small Language Models for Secure Context-Aware Instruction Following}

\author{
 \textbf{Kaiyan Zhang$^{1}$}, \textbf{Jianyu Wang$^{2}$}, \textbf{Ermo Hua}$^{1}$, \textbf{Biqing Qi$^{1,4}$}\\
 \textbf{Ning Ding$^{1,3}$}, \textbf{Bowen Zhou$^{1}$\Thanks{~Corresponding author}}
\\
$^1$ Tsinghua University, $^2$ Beijing Institute of Technology \\
$^3$ Frontis.AI, $^4$ Harbin Institute of Technology  \quad
\\
\texttt{zhang-ky22@mails.tsinghua.edu.cn}, \texttt{zhoubowen@tsinghua.edu.cn}
\\
\\
}

\begin{document}
\maketitle

\begin{abstract}

With the advancement of language models (LMs), their exposure to private data is increasingly inevitable, and their deployment (especially for smaller ones) on personal devices, such as PCs and smartphones, has become a prevailing trend.
In contexts laden with user information, enabling models to both safeguard user privacy and execute commands efficiently emerges as an essential research imperative.
In this paper, we propose CoGenesis, a collaborative generation framework integrating large (hosted on cloud infrastructure) and small models (deployed on local devices) to address privacy concerns logically.
Initially, we design a pipeline to create personalized writing instruction datasets enriched with extensive context details as the testbed of this research issue.
Subsequently, we introduce two variants of CoGenesis based on sketch and logits respectively.
Our experimental findings, based on our synthesized dataset and two additional open-source datasets, indicate that:
1) Large-scale models perform well when provided with user context but struggle in the absence of such context.
2) While specialized smaller models fine-tuned on the synthetic dataset show promise, they still lag behind their larger counterparts.
3) Our CoGenesis framework, utilizing mixed-scale models, showcases competitive performance, providing a feasible solution to privacy issues.
\end{abstract}

\section{Introduction}

\begin{figure}[ht]
  \centering
  \includegraphics[width=0.45\textwidth]{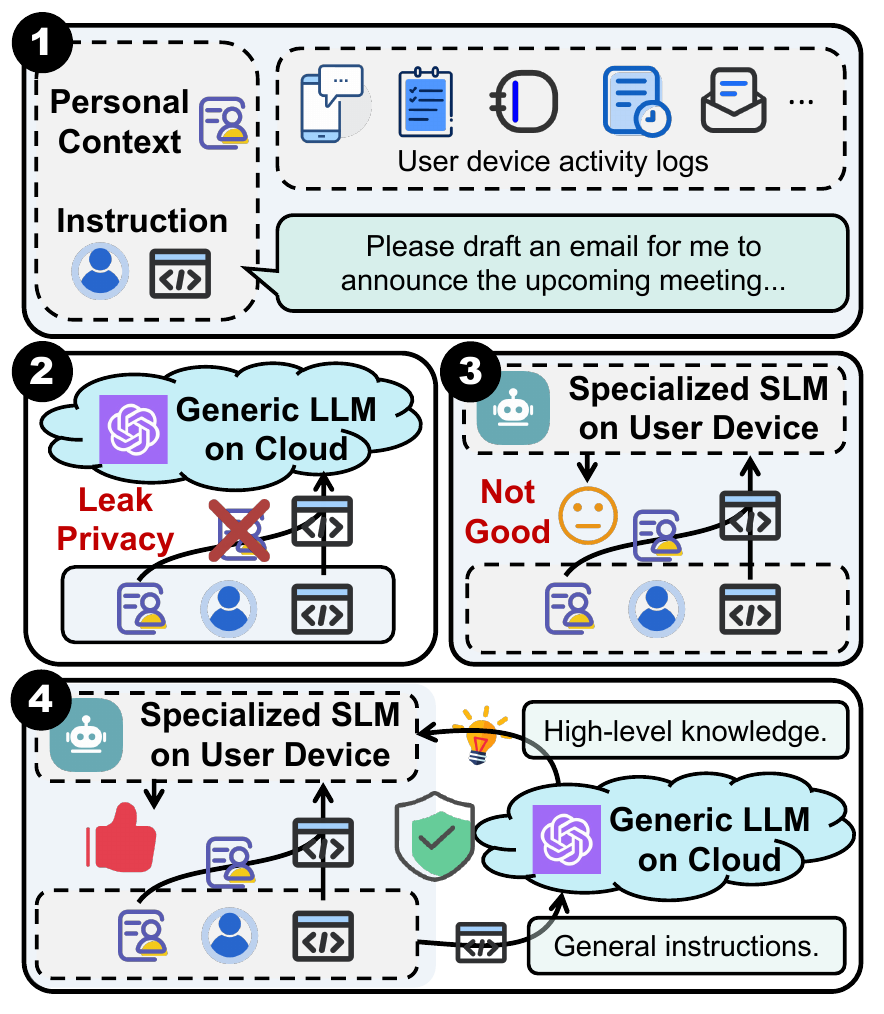} 
  \caption{\ding{202} Context-aware instruction following example. \ding{203} LLMs excel with context but risk privacy. \ding{204} Specialized and smaller LMs (SLMs) on device are privacy-friendly but underperform. \ding{205} Collaborating LLMs and SLMs enhances privacy and performance.}
  \label{fig:context_aware_example}
\end{figure}

Large Language Models (LLMs)\footnote{This paper defines large LMs (LLMs) as both closed and open-source models, designed for universal application and advanced performance, and intended for cloud deployment. Conversely, small LMs (SLMs) refer to models tailored for specific tasks and deployed on local devices.} 
have demonstrated significant potential in advancing artificial intelligence, exhibiting exceptional ability in instruction following and achieving superior performance in various tasks such as writing, coding, and other text-based activities~\citep{achiam2023gpt,bubeck2023sparks,touvron2023llama,touvron2023llama2}.
LLM-based AI assistants play a crucial role in executing instructions, aiding in writing tasks, and accelerating work processes, thereby fostering content innovation~\citep{zhang2023complete,haase2023artificial}.
LLMs often require extensive context information for generating more personalized and effective content, owing to their in-context learning abilities~\citep{brown2020language}.
Retrieval Augmented Generation (RAG)~\citep{gao2023retrieval} has proven beneficial in incorporating additional context to enhance the informativeness and personalization of LLM outputs.
However, current instruction-following tasks often fail to consider rich user context in their design~\citep{wang-etal-2023-self-instruct,xu2023baize,ding-etal-2023-enhancing}.
Incorporating personal experiences and activity logs could significantly augment the effectiveness of these instructions, called context-aware instruction following.

Despite their advancements, the most sophisticated LLMs, including GPT-4, Claude, and Gemini, 
are primarily commercialized and deployed on cloud services.
This API-based deployment ensures the privacy of the LLMs but potentially compromises user privacy~\citep{xiao2023offsite,zhang-etal-2023-crash}.
Although more powerful LLMs are being open-sourced, like Llama-2~\citep{touvron2023llama2}, Qwen~\citep{bai2023qwen}, and Mistral~\citep{jiang2024mixtral}, their stable deployment on local devices with limited resource remains challenging.
Recent advancements in smaller LMs (SLMs) equipped with billions of parameters now enable their deployment on consumer desktops and smartphones, achieving satisfactory performance~\citep{bai2023qwen,zhang2024tinyllama,singer2024h2o,minicpm2024}.
The balance between performance and privacy in LLMs and SLMs raises three critical questions: \textit{(1) How effectively can LLMs operate without stringent user privacy contexts?
(2) To what extent can specialized models, boasting billions of parameters, excel in context-aware instruction following?
(3) Is it possible to navigate the trade-off between performance and privacy through collaborations between large and small models?}

As indicated in Figure~\ref{fig:context_aware_example}, considering the following scenario in \ding{205}: smaller, personalized LMs are deployed on user devices with limited resources.
These SLMs can access private data and activity logs on the devices while processing instructions.
In contrast, the more advanced general LLMs operate on cloud services and receive only general instructions.
In this setup, the LLMs provide high-level knowledge like deeper planning, superior outlines, and even ``dark knowledge''.
Meanwhile, the SLMs utilize the context information and knowledge provided by the LLMs to collaboratively generate personalized content.
Current privacy protection methods for API-based services are limited~\citep{cummings2023challenges}; they are either capable of handling only simple classification tasks with text sanitization~\citep{kan2023protecting,chen2023customized} or require encryption or noise addition~\citep{zhou-etal-2022-textfusion,wu2023privacy}, which still pose a risk of data leakage.
In contrast, the collaborative generation approach involving SLMs and LLMs can logically prevent privacy breaches without the need to upload context information.

Overall, we highlight our contributions as follows:
(1) We introduce the context-aware instruction-following task that incorporates extensive user privacy context information. To support this, we design a four-step data construction process and synthesize a modest amount of instructional data for experimental validation.
(2) We investigate the performance of LLMs and SLMs on this task. Our findings indicate that SLMs, when provided with context, can outperform LLMs lacking context but lag behind the performance of LLMs equipped with context.
(3) For context-aware instruction generation, we present the CoGenesis framework. CoGenesis comprises sketch-based and logit-based variants to facilitate collaboration between large and small language models. This approach not only safeguards context privacy but also ensures performance gains.

\section{Context-aware Instruction Following}
\label{sec:context_aware}
\subsection{Task Definition}
\label{sec:context_aware_task}
Current instruction formats either consist of standalone instructions or instructions accompanied by additional inputs~\citep{wang-etal-2023-self-instruct}. While these instructions typically cover generic tasks such as writing, searching, and coding, the inputs often contain specific task information, such as tables and coding problems.
We classify the context-aware instruction-following task within the domain of controllable conditional text generation~\citep{zhang2023survey}, enriching standard instructions with additional personal context.
The task $\mathcal{T}$, focused on generating a response $r$ relevant to user-specific data marked by privacy and personal style, is directed by instructions $t$ and contextual information $\mathcal{C}$. It underscores the significance of contextual integration for enhancing output personalization and relevance, formally expressed as:
\begin{equation}
\mathcal{T}(t, \mathcal{C}): r = g(t, \mathcal{C}; \theta)
\end{equation}
This generative model $g$ with parameters $\theta$ aims to adaptively respond to user instructions within the nuanced context of individual data attributes.

\begin{figure*}[htbp]
  \centering
  \includegraphics[width=0.95\textwidth]{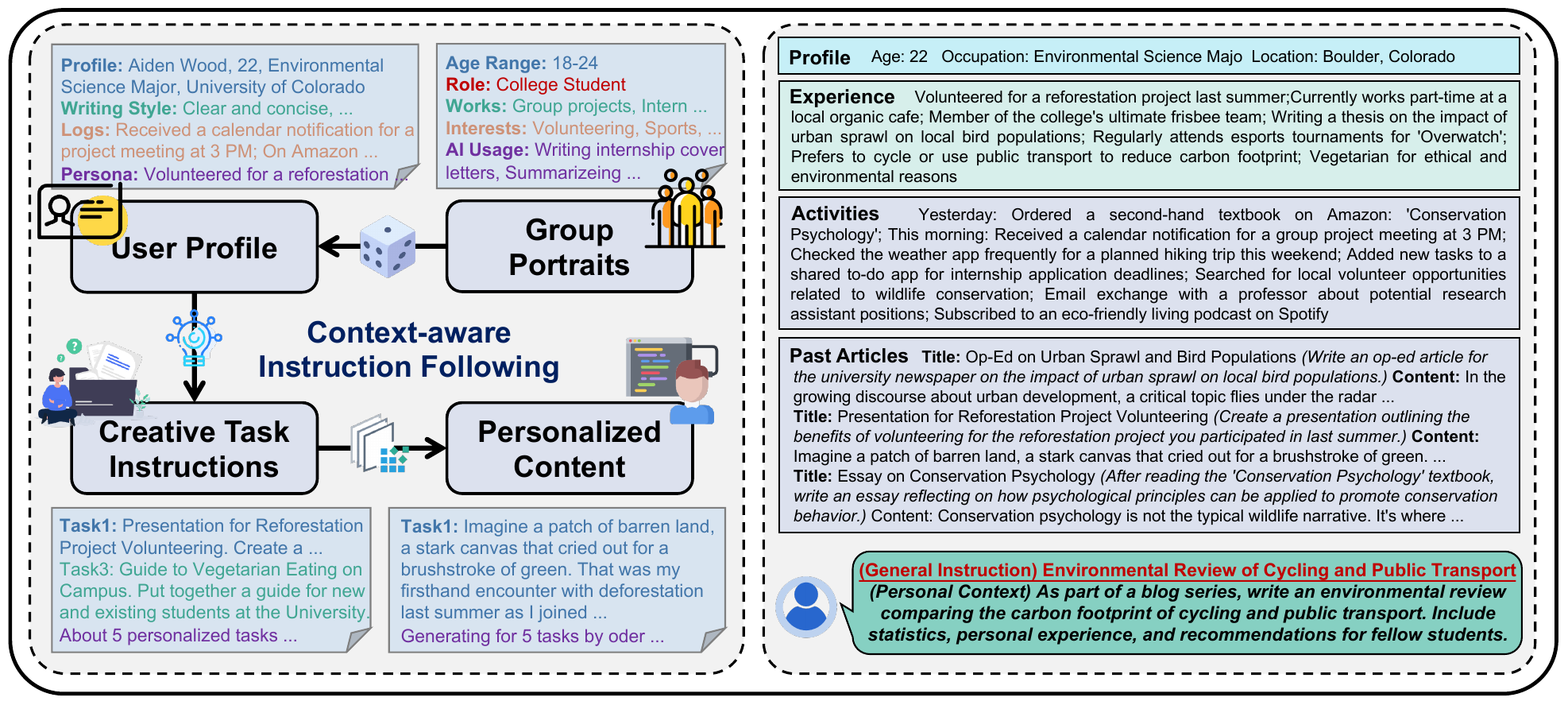} 
  \caption{This illustration demonstrates construction process and example of context-aware instructions.} 
  \label{fig:data_and_example}
\end{figure*}

\subsection{Data Construction}
\label{sec:context_aware_dataset}
As illustrated on the left side of Figure~\ref{fig:data_and_example}, we delineate a novel four-step pipeline for crafting context-aware instructions aimed at generating personalized and creative text with AI assistants for diverse user groups.
Our methodology begins with the creation of detailed user group portraits, capturing demographics, professional backgrounds, and interests to identify specific AI application scenarios. 
Individual user profiles are then elaborated, incorporating unique writing styles, fictional personal details, and smart device usage to construct nuanced characters for AI writing tasks.
These profiles inform the design of writing tasks that resonate with each character's lifestyle and digital interactions, ensuring task realism and relevance.
Finally, we generate personalized texts that reflect the characters' professional and personal narratives with stylistic accuracy, demonstrating our approach's efficacy in producing coherent, context-specific content for AI-facilitated text generation.

An example is illustrated on the right side of Figure~\ref{fig:data_and_example}, where the user instruction comprises a general section and a personal section. The latter, in conjunction with the provided profile, experience, activities, and previous articles, constitutes the privacy-sensitive context information.

\begin{figure*}[htbp]
  \centering
  \includegraphics[width=0.95\textwidth]{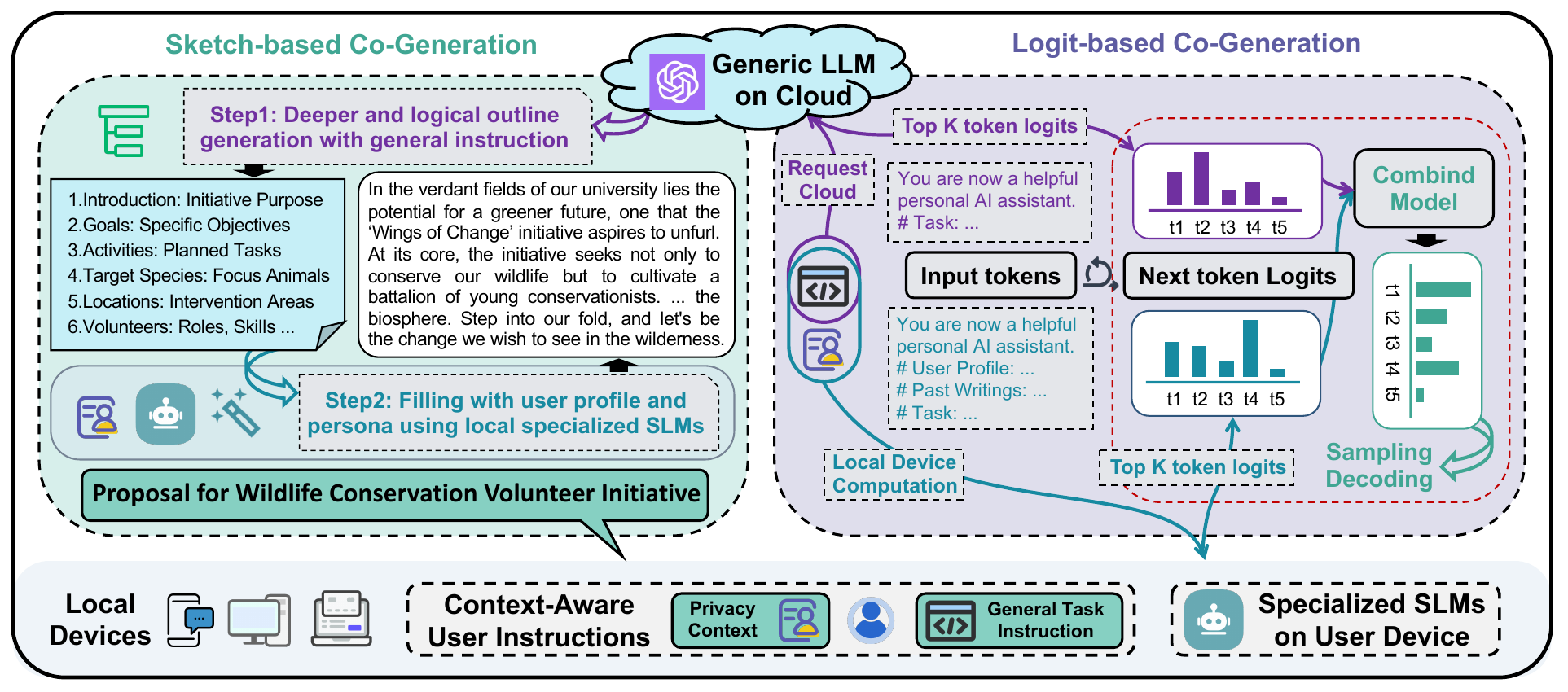} 
  \caption{This figure demonstrates two collaborative generation variants of CoGenesis framework.} 
  \label{fig:cogen_framework}
\end{figure*}

\section{Collaborative Generation Framework}
\label{sec:cogen}
\subsection{Overview of CoGenesis}
\label{sec:cogen_overview}
We present the CoGenesis framework that capitalizes on the strengths of two differently scaled models: a LLM with parameters $\theta_l$ and a SLM with parameters $\theta_s$.
This framework is centered around the fusion strategy, denoted as $f(\cdot)$, which intelligently combines the outputs from both models. 
Specifically, $\theta_l$ generates replies solely based on the general instruction $t$, while $\theta_s$ considers both user instruction $t$ and additional personal context $\mathcal{C}$ for its output generation.
The fusion strategy $f(\cdot)$ aims to synergistically blend the outputs of $\theta_l(r|t)$ and $\theta_s(r|t, \mathcal{C})$.
Intuitively, the combined performance is expected to not only surpass that of the individual models but also closely match the performance of $\theta_l$ had it processed both $r$ and $\mathcal{C}$.

In our collaborative framework, sketches (or outlines) of content and next token logits from LLMs are considered forms of high-level knowledge. The two approaches to the function $f(\cdot)$ are identified as sketch-based and logit-based, respectively. The sketch-based approach is model-agnostic, whereas the logit-based method requires LLMs and SLMs to share the same tokenizer in our present configuration. In the following sections, we will detail these two implementations of $f(\cdot)$ sequentially.

\subsection{Sketch-based CoGenesis}
\label{sec:cogen_sketch}
Recognizing the strengths of LLMs in planning and SLMs in crafting contextualized responses, we introduce a "sketch-then-fill" approach to synergize their capabilities for personalized content generation. 
As depicted on the left side of Figure~\ref{fig:cogen_framework}, this approach consists of two crucial steps:

\textbf{Step1: Sketch Generation by LLMs.}
Given the substantial cost and complexity, especially with API-dependent LLMs, we simplify the process by directly prompting LLMs with a general instruction $t$. The sketch $r_{\text{sketch}}$ of content is derived through text decoding from the LLMs using a sampling strategy, succinctly represented as:

\begin{equation}
r_{\text{sketch}} = \text{Decoding}_{\text{LLM}} (t; \theta_{l})
\end{equation}

\textbf{Step2: Content Personalization by SLMs.}
After acquiring the sketch $r_{\text{sketch}}$, the SLM utilizes it, along with the initial instruction $t$ and personal context $\mathcal{C}$, to tailor personalized content $r$. This phase focuses on fine-tuning the SLM's parameters $\theta_s$ to optimize content relevance and personalization:
\begin{equation}
\hat r =  \arg \max_r P (r | t, \mathcal{C}, r_{\text{sketch}}; \theta_s)
\end{equation}
In this formula, $r$ denotes the final, customized content, and $P$ is the likelihood of generating $r$ given instruction $t$, context $\mathcal{C}$, and sketch $r_{\text{sketch}}$ with parameters $\theta_s$.
This approach delineates the use of LLMs for foundational text sketching based on user prompts, without necessitating parameter adjustments, and SLMs for further content refinement. This ensures the final output aligns with user specifications and their interaction history, highlighting the distinct yet complementary roles of LLMs and SLMs in personalized content creation.

\subsection{Logit-based CoGenesis}
\label{sec:cogen_logit}
The logits produced in the final layer of language models encapsulate a wealth of information, reflecting the models' internal dark knowledge.
Previous efforts, such as contrastive decoding~\citep{li2022contrastive} and emulator tuning~\citep{mitchell2023emulator}, have explored the synergistic use of logits from both LLMs and SLMs to diminish hallucinations~\citep{sennrich2023mitigating}, augment reasoning capabilities~\citep{o2023contrastive}, and streamline the fine-tuning process of LLMs~\citep{liu2024tuning}.
Motivated by these works, our logit-based strategy involves integrating the logits of LLMs and SLMs under different inputs, ensuring collaborative determination of the subsequent token.
A notable aspect of our method is the differential context exposure for the models: SLMs access the full privacy context, while LLMs are provided with only broad instructions, as shown in Figure~\ref{fig:cogen_framework}.

Defining the response sequence up to the $k$th token as $r_{<k}$, and denoting the $k$th token probabilities over vocabulary generated by LLMs and SLMs as $p_{k}^{l}$ and $p_{k}^{s}$ respectively, we leverage a lightweight combined model, denoted as $\text{CombModel}$, with parameters $\theta_{c}$, to derive fusion weights $w$ for final combined probabilities $p_{k}^{c}$.
The computation of the logit-based method proceeds as follows:
\begin{equation}
    \label{eq:logit_pks}
    p_{k}^{s} = \theta_s(r_{<k}, t, \mathcal{C}),\ p_{k}^{l} = \theta_l(r_{<k}, t)
\end{equation}
\begin{equation}
    \label{eq:logit_weights}
    w = \text{CombModel}(p_{k}^l, p_{k}^s)
\end{equation}
\begin{equation}
    \label{eq:logit_pkc}
    p_{k}^{c} = w \cdot p_{k}^{s} + (1-w) \cdot p_{k}^{l}
\end{equation}

\section{Experiments}

\subsection{Datasets}
\textbf{Synthetic Dataset.}
Following the construction process outlined in $\S$~\ref{sec:context_aware_dataset}, we synthesize a context-aware instruction-following dataset with GPT-4. Due to cost considerations, we construct a dataset representing thousands of fictitious users, based on hundreds of group portraits. After conducting quality and format filtering, we obtained a total of 1,500 users for training and validation. Additionally, we selected approximately 200 users from diverse group portraits to serve as the test set.

\textbf{Open-source Datasets.}
In addition to our synthesized context-aware instruction datasets, we also utilize publicly accessible, personalized context writing datasets, although they are limited to specific tasks in domains such as email and academic papers.
Specifically, we employ the processed Avocado Research Email and Citation Network Papers datasets in LaMP~\citep{salemi2023lamp}.
Furthermore, we refine these datasets to facilitate the generation of email bodies and paper abstracts, considering previous emails and papers from the same users as contextual information.

Further details about the synthesized and processed datasets can be found in Appendix~\ref{apx:dataset_details}.

\subsection{Baselines}

\textbf{Settings.} We primarily evaluate four configurations in our experiments:
\textbf{1) \textit{LLM with context.}} We engage LLMs with additional context to facilitate personalized generation. It's an \textit{upper bound} that may compromise context privacy.
\textbf{2) \textit{LLM w/o context.}} Given the importance of privacy in contextual data, it is advisable to limit requests to cloud-based LLMs to general instructions only. This setting serves as a \textit{lower bound} that preserves user privacy.
\textbf{3) \textit{SLM with context.}} This setting establishes the baseline for privacy-protected, on-device personalized generation.
\textbf{4) \textit{SLM + LLM with context.}} This is our proposed collaborative generation between large- and small-scale models. Within this framework, we evaluate sketch-based and logit-based methods.

\textbf{Models.} Our selection of LLMs encompasses both commercial API-based and open-source models. For the commercial segment, we concentrate on GPT-3.5-turbo and GPT-4-turbo. In the realm of open-source, we opt for the largest models within the Llama-2~\citep{touvron2023llama2}, Qwen (72B in versions 1 and 1.5~\footnote{\url{https://github.com/QwenLM/Qwen1.5}})~\citep{bai2023qwen}, and Mistral~\citep{jiang2024mixtral} series.
Regarding SLMs, we prioritize the most recently released models with 1$\sim$2 billions of parameters. This includes TinyLlama~\citep{zhang2024tinyllama}, Qwen (1.8B in versions 1 and 1.5), StableLM~\footnote{\url{https://huggingface.co/stabilityai}}, and H2O-Danube~\citep{singer2024h2o}.
For both LLMs and SLMs, our focus is on the chat versions, employing the default template for each model for consistency.

Additionally, zero-shot LLMs are used in both \textit{with context} and \textit{w/o context} settings. For \textit{SLM with context}, we include both zero-shot and fine-tuned models, whereas only fine-tuned SLMs are utilized in \textit{SLM + LLM with context}. 
Further information on prompts for LLMs and fine-tuning details for SLMs is presented in Appendix~\ref{apx:model_details}.

\begin{table*}[ht]
  \small
  \centering
  \begin{adjustbox}{width=\textwidth}
  \begin{tabular}{lllllllllll}
    \toprule
    \multirow{2}{*}{Model} & \multirow{2}{*}{Params} & \multicolumn{3}{c}{Context-aware Instructions} & \multicolumn{3}{c}{Avocado Emails} & \multicolumn{3}{c}{Academic Paper Abstracts} \\
    \cmidrule(lr){3-5} \cmidrule(lr){6-8} \cmidrule(lr){9-11} 
                        &       & Ovl.(w) & Per. & Ovl.(w/o) & Ovl.(w) & Per. & Ovl.(w/o) & Ovl.(w) & Per. &  Ovl.(w/o)  \\
    \midrule
    \multicolumn{11}{l}{ \quad \quad \textbf{\textit{zero-shot LLM with context (upper bound)}}} \\
     L1 GPT-4-turbo     & N/A   &  \textbf{8.85}  &  \textbf{8.90}  &  \textbf{8.54}  &  \textbf{8.31}  &  \textbf{7.71}  &  \textbf{8.05}  &  \textbf{8.64}  &  \textbf{8.47}  &  \textbf{8.68} \\
     L2 GPT-3.5-turbo   & N/A   &  8.30  &  8.33  &  7.58  &  7.70  &  7.45  &  7.57  &  7.97  &  7.88  &  8.29 \\
     L3 Llama-2-Chat    & 70B   &  7.78  &  7.98  &  7.96  &  7.00  &  7.18  &  \underline{8.05}  &  7.48  &  7.12  &  8.32 \\
     L4 Qwen-Chat(v1)   & 72B   &  8.38  &  8.38  &  8.14  &  7.62  &  7.10  &  7.70  &  7.90  &  7.62  &  8.24 \\
     L5 Qwen-Chat(v1.5) & 72B   &  \underline{8.70}  &  \underline{8.67}  &  \underline{8.26}  &  \underline{8.20}  &  \underline{7.69}  &  7.80  &  \underline{8.52}  &  \underline{8.16}  &  \underline{8.60} \\
     L6 Mixtral-8x7b    & 47B   &  8.12  &  8.22  &  7.96  &  7.35  &  6.88  &  6.92  &  7.92  &  7.62  &  8.08 \\
     \midrule
     \midrule
     \multicolumn{11}{l}{ \quad \quad \textbf{\textit{zero-shot LLM w/o context (lower bound)}}} \\
     L1 GPT-4-turbo     & N/A   &  \textbf{6.10}  &  \textbf{6.46}  &  \textbf{8.75}  &  \underline{5.10}  &  \underline{3.79}  &  \textbf{8.58}  &  \textbf{8.42}  &  \textbf{7.94}  &  \textbf{8.73} \\
     L2 GPT-3.5-turbo   & N/A   &  4.34  &  3.76  &  7.47  &  3.72  &  3.03  &  7.61  &  3.37  &  3.59  &  6.04 \\
     L3 Llama-2-Chat    & 70B   &  4.74  &  4.60  &  8.12  &  3.38  &  3.52  &  8.04  &  6.74  &  6.18  &  7.90 \\
     L4 Qwen-Chat(v1)   & 72B   &  3.70  &  3.38  &  7.72  &  3.28  &  2.38  &  7.50  &  6.42  &  5.54  &  7.90 \\
     L5 Qwen-Chat(v1.5) & 72B   &  \underline{5.86}  &  \underline{5.98}  &  \underline{8.52}  &  \textbf{5.83}  &  \textbf{4.40}  &  \underline{8.54}  &  \underline{7.92}  &  \underline{7.16}  &  \underline{8.28} \\
     L6 Mixtral-8x7b    & 47B   &  5.32  &  5.08  &  8.14  &  3.38  &  3.52  &  8.04  &  6.74  &  6.18  &  7.90 \\
     \midrule
     \midrule
     \multicolumn{11}{l}{ \quad \quad \textbf{\textit{zero-shot SLM with context}}} \\
     S1 StableLM-Zephyr &  1.6B &  \textbf{6.88}  &  6.82  &  \underline{6.68}  &  \textbf{6.03}  &  \textbf{5.49}  &  \underline{6.51}  &  \textbf{7.32}  &  \textbf{6.94}  &  \textbf{8.14} \\
     S2 H2O-danube-chat &  1.8B &  6.56  &  \textbf{6.94}  &  6.60  &  4.77  &  \underline{5.00}  &  \textbf{6.57}  &  5.98  &  5.54  &  6.94 \\
     S3 TinyLlama-Chat  &  1.1B &  1.72  &  1.84  &  2.14  &  4.00  &  3.54  &  5.40  &  1.96  &  1.88  &  4.24 \\
     S4 Qwen-Chat(v1)   &  1.8B &  5.78  &  5.50  &  6.00  &  4.91  &  4.54  &  6.49  &  4.46  &  5.04  &  6.74 \\
     S5 Qwen-Chat(v1.5) &  1.8B &  \underline{6.86}  &  \underline{6.86}  &  \textbf{7.20}  &  \underline{5.51}  &  4.89  &  6.14  &  \underline{6.42}  &  \underline{6.14}  &  \underline{7.78} \\
     \midrule
     \midrule
     \multicolumn{11}{l}{ \quad \quad \textbf{\textit{finetuned SLM (+ LLM) with context}}} \\
     S1 StableLM-Zephyr 
                    &  1.6B &  8.30 &  8.56  &  7.73  &  7.58  &  6.70  &  7.20  &  7.96  &  7.64  &  8.18 \\
     \quad \quad \textit{+ L1 sketch}
                    & \textit{mixed}  &  8.48\textcolor{red}{$_{\uparrow 0.18}$} &  8.56\textcolor{red}{$_{\uparrow 0.00}$}  &  7.98\textcolor{red}{$_{\uparrow 0.25}$}  &  7.68\textcolor{red}{$_{\uparrow 0.10}$}  &  6.62\textcolor{blue}{$_{\downarrow 0.08}$}  &  7.48\textcolor{red}{$_{\uparrow 0.28}$}  &  8.28\textcolor{red}{$_{\uparrow 0.32}$}  &  7.48\textcolor{blue}{$_{\downarrow 0.16}$}  &  8.38\textcolor{red}{$_{\uparrow 0.20}$} \\ \midrule
     S2 H2O-danube-chat 
                    &  1.8B &  7.64 &  7.58  &  7.00  &  6.50  &  6.16  &  6.34  &  7.70  &  7.30  &  8.06 \\
     \quad \quad \textit{+ L1 sketch}
                    & \textit{mixed}  &  7.84\textcolor{red}{$_{\uparrow 0.20}$} &  7.78\textcolor{red}{$_{\uparrow 0.20}$}  &  7.14\textcolor{red}{$_{\uparrow 0.14}$}  &  7.14\textcolor{red}{$_{\uparrow 0.64}$}  &  6.72\textcolor{red}{$_{\uparrow 0.56}$}  &  7.52\textcolor{red}{$_{\uparrow 1.18}$}  &  8.10\textcolor{red}{$_{\uparrow 0.40}$}  &  7.28\textcolor{blue}{$_{\downarrow 0.02}$}  &  8.18\textcolor{red}{$_{\uparrow 0.12}$} \\ \midrule
     S3 TinyLlama-Chat  
                    &  1.1B &  7.42 &  7.66  &  6.78  &  6.12  &  5.92  &  6.20  &  7.66  &  7.32  &  8.18 \\
     \quad \quad \textit{+ L1 sketch}
                    & \textit{mixed}  &  7.66\textcolor{red}{$_{\uparrow 0.24}$} &  7.14\textcolor{blue}{$_{\downarrow 0.52}$}  &  6.82\textcolor{red}{$_{\uparrow 0.04}$}  &  6.58\textcolor{red}{$_{\uparrow 0.46}$}  &  6.02\textcolor{red}{$_{\uparrow 0.10}$}  &  6.60\textcolor{red}{$_{\uparrow 0.40}$}  &  7.72\textcolor{red}{$_{\uparrow 0.06}$}  &  7.36\textcolor{red}{$_{\uparrow 0.04}$}  &  8.10\textcolor{blue}{$_{\downarrow 0.08}$} \\
     \quad \quad \textit{+ L3 logits}
                    & \textit{mixed}  &  7.76\textcolor{red}{$_{\uparrow 0.34}$}  &  7.74\textcolor{red}{$_{\uparrow 0.08}$}  &  7.06\textcolor{red}{$_{\uparrow 0.28}$}  &   6.06\textcolor{blue}{$_{\downarrow 0.06}$}  &  6.16\textcolor{red}{$_{\uparrow 0.24}$}  &  6.94\textcolor{red}{$_{\uparrow 0.74}$}  &  8.14\textcolor{red}{$_{\uparrow 0.48}$}  &  7.34\textcolor{red}{$_{\uparrow 0.02}$}   &  8.04\textcolor{blue}{$_{\downarrow 0.14}$}  \\ \midrule
     S4 Qwen-Chat(v1)   
                    &  1.8B &  7.44 &  7.76  &  7.02  &  7.00  &  6.71  &  7.06  &  7.84  &  7.36  &  8.18 \\
     \quad \quad \textit{+ L1 sketch}
                    & \textit{mixed}  &  7.80\textcolor{red}{$_{\uparrow 0.36}$} &  7.82\textcolor{red}{$_{\uparrow 0.06}$}  &  7.64\textcolor{red}{$_{\uparrow 0.62}$}  &  7.18\textcolor{red}{$_{\uparrow 0.18}$}  &  6.44\textcolor{blue}{$_{\downarrow 0.27}$}  &  7.28\textcolor{red}{$_{\uparrow 0.22}$}  &  8.02\textcolor{red}{$_{\uparrow 0.18}$}  &  7.70\textcolor{red}{$_{\uparrow 0.34}$}  &  8.34\textcolor{red}{$_{\uparrow 0.16}$} \\
     \quad \quad \textit{+ L4 logits}
                    & \textit{mixed}  &  8.12\textcolor{red}{$_{\uparrow 0.68}$} &  8.20\textcolor{red}{$_{\uparrow 0.44}$}  &  7.86\textcolor{red}{$_{\uparrow 0.84}$}  &  7.48\textcolor{red}{$_{\uparrow 0.48}$}  &  6.44\textcolor{blue}{$_{\downarrow 0.27}$}  &  7.46\textcolor{red}{$_{\uparrow 0.40}$}  &  7.92\textcolor{red}{$_{\uparrow 0.08}$}  &  7.16\textcolor{blue}{$_{\downarrow 0.20}$}      &  8.30\textcolor{red}{$_{\uparrow 0.12}$}  \\ \midrule
     S5 Qwen-Chat(v1.5)
                    &  1.8B &  8.08 &  8.12  &  7.40  &  6.34  &  5.56  &  6.54  &  7.68  &  7.30  &  8.18 \\
     \quad \quad \textit{+ L1 sketch}
                    & \textit{mixed}  &  8.18\textcolor{red}{$_{\uparrow 0.10}$} &  7.98\textcolor{blue}{$_{\downarrow 0.14}$}  &  7.62\textcolor{red}{$_{\uparrow 0.22}$}  &  6.54\textcolor{red}{$_{\uparrow 0.20}$}  &  5.84\textcolor{red}{$_{\uparrow 0.28}$}  &  6.74\textcolor{red}{$_{\uparrow 0.20}$}  &  8.10\textcolor{red}{$_{\uparrow 0.42}$}  &  7.24\textcolor{blue}{$_{\downarrow 0.06}$}  &  8.32\textcolor{red}{$_{\uparrow 0.14}$} \\
     \quad \quad \textit{+ L5 logits}
                    & \textit{mixed} & 8.28\textcolor{red}{$_{\uparrow 0.20}$}  &  8.22\textcolor{red}{$_{\uparrow 0.10}$}  &  7.80\textcolor{red}{$_{\uparrow 0.40}$}  &   6.66\textcolor{red}{$_{\uparrow 0.32}$} & 5.70\textcolor{red}{$_{\uparrow 0.14}$} &  7.12\textcolor{red}{$_{\uparrow 0.58}$} & 8.14\textcolor{red}{$_{\uparrow 0.46}$} & 7.28\textcolor{blue}{$_{\downarrow 0.02}$} & 8.40\textcolor{red}{$_{\uparrow 0.22}$} \\
    \bottomrule
  \end{tabular}
  \end{adjustbox}
  \caption{
    The table displays the performance of LLMs and SLMs in both \textit{with context} and \textit{w/o context} settings. We highlight the best result in \textbf{bold}, the second in \underline{underline} and indicate variations in each SLM using \textcolor{red}{$\uparrow$} and \textcolor{blue}{$\downarrow$}.
  }
  \label{tab:main_results}
\end{table*}

\subsection{Evaluation Metrics}
For instruction-following evaluation, LLM-based evaluators like GPT-4 have shown high consistency and effectiveness~\citep{chang2023survey}, particularly in the generation of personalized content, outperforming human evaluators in consistency~\citep{wang2023automated}.
Therefore, we employ GPT-4-turbo as a judge to assess generated content from multiple perspectives.
This evaluation encompasses several criteria: personalization, alignment with the user profile, helpfulness, relevance, depth, creativity, and the level of detail, collectively contributing to the overall score (\textbf{Ovl.}). To further analyze the performance, we introduce two distinct prompts: one incorporating user context (\textbf{Ovl.(w)}) and the other excluding it (\textbf{Ovl.(w/o)}), enabling a comparative assessment of LLMs in both personalized content generation and broad instruction adherence. Additionally, the personalized scores of the responses are assessed independently, indicated as \textbf{Per.}

\subsection{Main Results}
As illustrated in Table~\ref{tab:main_results}, we analyze the experimental results considering the following aspects:

\textbf{Results on LLMs.}
In the \textit{with context} setting, \texttt{GPT-4-turbo} outperformed all other models across all metrics and datasets, followed by \texttt{Qwen-Chat(v1.5)} with the second-best performance. Similar outcomes were observed in the \textit{w/o context} setting.
However, a comparison between the two settings reveals that all LLMs exhibit diminished performance in terms of personalization and overall scores in the absence of context information, underscoring the value of context.
Interestingly, while context greatly influences the scores in \textit{with context} setting, it has a negligible effect on scores in \textit{w/o context} setting, reflecting advanced capability of LLMs in adhering to general instructions.
Moreover, when comparing our synthesized context-aware dataset and the email dataset to the paper abstract dataset, the latter demonstrates limited personalization factors resulting in high performance in \textit{w/o context} setting.

\begin{figure*}[t]
  \centering
  \includegraphics[width=0.90\textwidth]{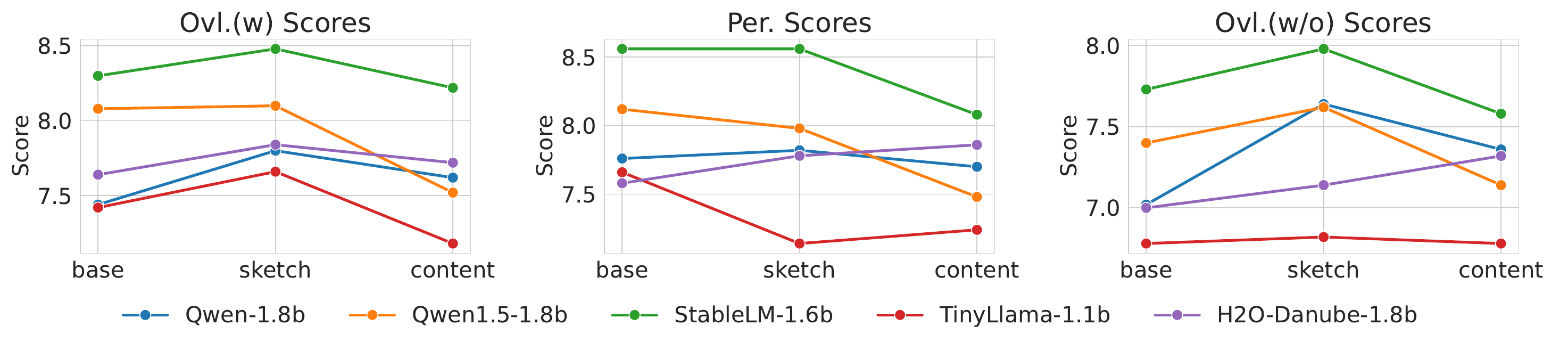} 
  \caption{Comparative Performance of SLMs Utilizing Sketch versus Full Content.}
\label{fig:ab_sketch_content}
\end{figure*}

\begin{figure}[t]
  \centering
  \includegraphics[width=0.5\textwidth]{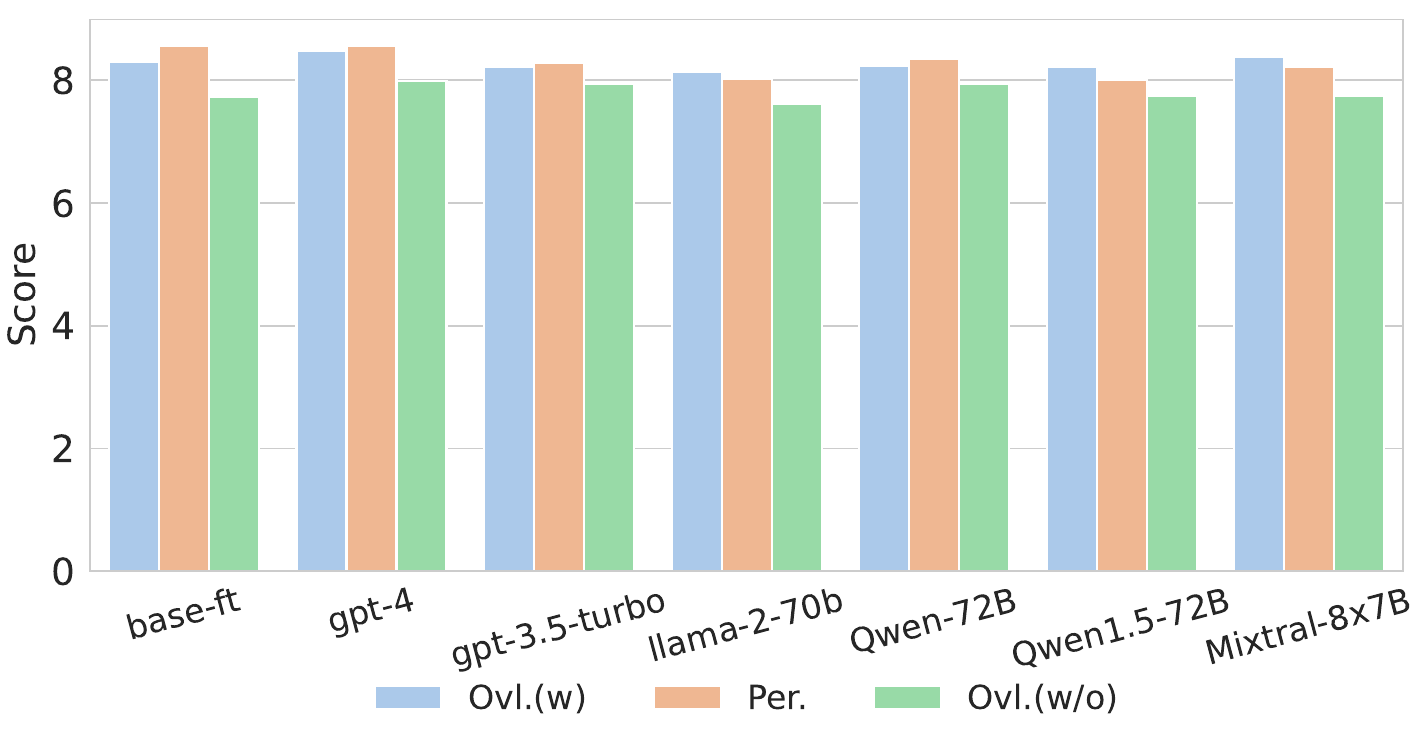} 
  \caption{Performance of StableLM Using Sketches Generated by Various Models.}
\label{fig:ab_sketch_across}
\end{figure}

\textbf{Results on SLMs.}
Zero-shot SLMs exhibit highly varied performances, influenced by pre-training and supervised fine-tuning factors.
Notably, \texttt{StableLM-Zephyr} achieves the highest performance, with \texttt{H2O-Danube-Chat} and \texttt{Qwen-Chat(v1.5)} closely competing for second place.
Interestingly, with the advantage of context, zero-shot SLMs can outperform LLMs in scenarios lacking context.
However, LLMs consistently outperform SLMs in overall scores without context, underscoring their superior capabilities in general instruction generation.
After fine-tuning, SLMs surpass the performance of many LLMs with context, demonstrating the benefits of specialization.
Yet, SLMs do not reach the performance levels of the most powerful LLMs, such as \texttt{GPT-4-turbo} and \texttt{Qwen-Chat(v1.5)}. These results highlight the promise of collaboration between specialized SLMs and LLMs for achieving better personalized scores, deeper writing, and enhanced instruction generalization.

\textbf{Results on Mixed-Scale Models Collaborations.}
In our exploration of mixed-scale model collaboration, we utilize the sketch-based method for all SLMs and the logit-based method exclusively within the Llama and Qwen model families.
This comparison reveals that collaborations between mixed-scale models achieve results comparable to those of LLMs alone, while also safeguarding privacy. 
Collaborative efforts generally enhance the overall scores of SLMs, both in evaluations with and without user context.
Nevertheless, incorporating sketch-based collaboration, in particular, might slightly detract from personalization scores due to the reliance on LLMs that inference without context.
Between the logit-based and sketch-based approaches, the former proves more efficacious, contingent upon the SLMs and LLMs utilizing a common tokenizer.
Overall, this collaborative strategy between mixed-scale models offers a promising avenue for balancing efficiency and privacy considerations, though it still necessitates further refinements to optimize performance.

\subsection{Ablation Study}

\subsubsection{Sketch-based CoGenesis}

\textbf{Sketch vs. Full Content.}
Figure~\ref{fig:ab_sketch_content} contrasts the efficacy of employing merely the sketch versus the entire content provided by LLMs.
The findings indicate that incorporating full content generally detracts from model performance, in contrast to utilizing no content or only the sketch.
This discrepancy can be attributed to the potential overload of redundant information.
As explored in~\citep{weston2023system}, an excess of content does not invariably enhance performance and risks the inclusion of extraneous details.

\textbf{Generalization Capabilities of Sketch.}
For small models fine-tuned with sketches generated by GPT-4, we explore their generalization potential by utilizing sketches from a variety of LLMs during testing. Specifically, we focus on \texttt{StableLM-Zephyr} for detailed ablation analysis. Figure~\ref{fig:ab_sketch_across} demonstrates that employing sketches from alternative LLMs marginally affects the overall and personalized scores negatively but enhances the overall score in scenarios lacking context, relative to \texttt{GPT-4}. This suggests that sketches generated by different models vary and exhibit limited generalization capabilities.

\begin{table}[t]
\centering
\small
\scalebox{0.9}{
\begin{tabular}{c|ccc}
\toprule
 \diagbox{Models}{Metric}  & Ovl.(w) & Per. & Ovl.(w/o) \\
 \midrule
    Qwen-72B-Chat (\textit{with context}) & \textbf{8.38}  &  \textbf{8.38}  &  \textbf{8.14}  \\
    Qwen-72B-Chat (\textit{w/o context})& 3.70  &  3.38  &  7.72 \\
    Qwen-1.8B-Chat (\textit{finetuned})    & 7.44  &  7.76 &  7.02 \\ \midrule
    Mean Pooling Fusing          & 7.76  &  7.84 &  7.42 \\
    Max Pooling Fusing           & 7.90  &  7.94 &  7.52 \\
    Learnable Weights Fusing     & \textbf{8.12}  &  \textbf{8.20} &  \textbf{7.86} \\
\bottomrule
\end{tabular}}
\caption{Comparison of Mean and Max Fusion Strategies Against our Learnable Fusion Model.}
\label{tab:ab_logits_fusion}
\end{table}

\subsubsection{Logit-based CoGenesis}
\begin{figure*}[htbp]
  \centering
  \fbox{
  \begin{minipage}{1.0\textwidth}
    \small
 \begin{CJK*}{UTF8}{gbsn}
{\setlength{\fboxsep}{0pt}\colorbox{white!0}{\parbox{1.0\textwidth}{
\colorbox{red!50.0}{\strut Subject} \colorbox{blue!16.0}{\strut :} \colorbox{red!6.0}{\strut  Join} \colorbox{blue!4.0}{\strut  Us} \colorbox{blue!2.0}{\strut  for} \colorbox{blue!46.0}{\strut  a} \colorbox{red!10.0}{\strut  Day} \colorbox{blue!68.0}{\strut  of} \colorbox{blue!4.0}{\strut  Giving} \colorbox{red!14.000000000000002}{\strut  at} \colorbox{blue!18.0}{\strut  Homes} \colorbox{blue!30.0}{\strut  for} \colorbox{red!20.0}{\strut  Humanity} \colorbox{red!22.0}{\strut  -} \colorbox{red!4.0}{\strut  Your} \colorbox{red!4.0}{\strut  Hands} \colorbox{blue!36.0}{\strut ,} \colorbox{blue!10.0}{\strut  Our} \colorbox{blue!14.000000000000002}{\strut  Community} \colorbox{red!16.0}{\strut !

} \colorbox{red!34.0}{\strut Dear} \colorbox{red!46.0}{\strut  [} \colorbox{red!4.0}{\strut Recipient} \colorbox{blue!68.0}{\strut 's} \colorbox{blue!12.0}{\strut  Name} \colorbox{blue!74.0}{\strut ],

} \colorbox{red!60.0}{\strut As} \colorbox{blue!8.0}{\strut  a} \colorbox{red!20.0}{\strut  dedicated} \colorbox{red!16.0}{\strut  volunteer} \colorbox{blue!36.0}{\strut  for} \colorbox{red!14.000000000000002}{\strut  Homes} \colorbox{blue!18.0}{\strut  for} \colorbox{red!16.0}{\strut  Humanity} \colorbox{blue!20.0}{\strut ,} \colorbox{blue!40.0}{\strut  I} \colorbox{blue!22.0}{\strut 've} \colorbox{blue!2.0}{\strut  had} \colorbox{blue!60.0}{\strut  the} \colorbox{blue!10.0}{\strut  privilege} \colorbox{blue!82.0}{\strut  of} \colorbox{blue!30.0}{\strut  witnessing} \colorbox{blue!48.0}{\strut  firsthand} \colorbox{blue!84.0}{\strut  the} \colorbox{red!2.0}{\strut  transformative} \colorbox{blue!18.0}{\strut  power} \colorbox{blue!78.0}{\strut  of} \colorbox{blue!20.0}{\strut  community} \colorbox{blue!32.0}{\strut  spirit} \colorbox{blue!38.0}{\strut .} \colorbox{blue!0.0}{\strut  It} \colorbox{blue!40.0}{\strut 's} \colorbox{blue!52.0}{\strut  a} \colorbox{blue!14.000000000000002}{\strut  testament} \colorbox{blue!76.0}{\strut  to} \colorbox{blue!80.0}{\strut  the} \colorbox{blue!38.0}{\strut  human} \colorbox{blue!40.0}{\strut  spirit} \colorbox{blue!60.0}{\strut  that} \colorbox{blue!68.0}{\strut  we} \colorbox{blue!72.0}{\strut  can} \colorbox{blue!34.0}{\strut  turn} \colorbox{blue!52.0}{\strut  a} \colorbox{blue!10.0}{\strut  simple} \colorbox{blue!18.0}{\strut  act} \colorbox{blue!52.0}{\strut  of} \colorbox{blue!24.0}{\strut  kindness} \colorbox{blue!40.0}{\strut  into} \colorbox{blue!68.0}{\strut  a} \colorbox{blue!8.0}{\strut  beacon} \colorbox{blue!92.0}{\strut  of} \colorbox{blue!60.0}{\strut  hope} \colorbox{blue!54.0}{\strut  for} \colorbox{blue!18.0}{\strut  those} \colorbox{blue!84.0}{\strut  in} \colorbox{blue!16.0}{\strut  need} \colorbox{red!4.0}{\strut .} \colorbox{red!4.0}{\strut  This} \colorbox{red!2.0}{\strut  month} \colorbox{blue!18.0}{\strut ,} \colorbox{blue!50.0}{\strut  I} \colorbox{blue!20.0}{\strut  invite} \colorbox{blue!14.000000000000002}{\strut  you} \colorbox{blue!68.0}{\strut  to} \colorbox{blue!40.0}{\strut  join} \colorbox{blue!64.0}{\strut  me} \colorbox{blue!44.0}{\strut  in} \colorbox{blue!56.00000000000001}{\strut  celebrating} \colorbox{blue!68.0}{\strut  this} \colorbox{blue!20.0}{\strut  spirit} \colorbox{blue!56.00000000000001}{\strut  of} \colorbox{blue!16.0}{\strut  giving} \colorbox{blue!28.000000000000004}{\strut  at} \colorbox{blue!22.0}{\strut  our} \colorbox{red!4.0}{\strut  upcoming} \colorbox{blue!16.0}{\strut  charity} \colorbox{red!4.0}{\strut  event} \colorbox{red!12.0}{\strut .

} \colorbox{red!28.000000000000004}{\strut Date} \colorbox{red!16.0}{\strut :} \colorbox{red!4.0}{\strut  [} \colorbox{red!4.0}{\strut Event} \colorbox{red!12.0}{\strut  Date} \colorbox{red!2.0}{\strut ]
} \colorbox{blue!0.0}{\strut Time} \colorbox{red!22.0}{\strut :} \colorbox{red!38.0}{\strut  [} \colorbox{red!50.0}{\strut Event} \colorbox{blue!10.0}{\strut  Time} \colorbox{blue!4.0}{\strut ]
} \colorbox{blue!8.0}{\strut Location} \colorbox{red!24.0}{\strut :} \colorbox{red!64.0}{\strut  [} \colorbox{red!8.0}{\strut Event} \colorbox{blue!30.0}{\strut  Venue} \colorbox{blue!16.0}{\strut ]

} \colorbox{red!22.0}{\strut At} \colorbox{blue!48.0}{\strut  this} \colorbox{red!4.0}{\strut  event} \colorbox{blue!18.0}{\strut ,} \colorbox{blue!54.0}{\strut  we} \colorbox{red!2.0}{\strut 'll} \colorbox{blue!28.000000000000004}{\strut  be} \colorbox{blue!6.0}{\strut  coming} \colorbox{blue!34.0}{\strut  together} \colorbox{blue!64.0}{\strut  to} \colorbox{blue!18.0}{\strut  build} \colorbox{blue!46.0}{\strut  not} \colorbox{blue!60.0}{\strut  just} \colorbox{blue!62.0}{\strut  homes} \colorbox{blue!44.0}{\strut ,} \colorbox{blue!22.0}{\strut  but} \colorbox{blue!62.0}{\strut  futures} \colorbox{blue!34.0}{\strut .} \colorbox{blue!22.0}{\strut  With} \colorbox{blue!66.0}{\strut  each} \colorbox{blue!34.0}{\strut  hammer} \colorbox{blue!48.0}{\strut  strike} \colorbox{blue!48.0}{\strut  and} \colorbox{blue!28.000000000000004}{\strut  each} \colorbox{blue!24.0}{\strut  plank} \colorbox{blue!26.0}{\strut  placed} \colorbox{blue!44.0}{\strut ,} \colorbox{blue!80.0}{\strut  we} \colorbox{blue!76.0}{\strut 'll} \colorbox{blue!48.0}{\strut  be} \colorbox{blue!30.0}{\strut  crafting} \colorbox{blue!64.0}{\strut  not} \colorbox{blue!52.0}{\strut  just} \colorbox{blue!74.0}{\strut  shelter} \colorbox{blue!56.00000000000001}{\strut ,} \colorbox{blue!38.0}{\strut  but} \colorbox{blue!68.0}{\strut  a} \colorbox{blue!56.00000000000001}{\strut  sense} \colorbox{blue!94.0}{\strut  of} \colorbox{blue!16.0}{\strut  belonging} \colorbox{blue!50.0}{\strut  and} \colorbox{blue!18.0}{\strut  stability} \colorbox{blue!50.0}{\strut  for} \colorbox{blue!28.000000000000004}{\strut  families} \colorbox{blue!80.0}{\strut  in} \colorbox{red!2.0}{\strut  our} \colorbox{red!6.0}{\strut  community} \colorbox{blue!56.00000000000001}{\strut .} \colorbox{red!6.0}{\strut  It} \colorbox{blue!57.99999999999999}{\strut 's} \colorbox{blue!62.0}{\strut  a} \colorbox{blue!8.0}{\strut  chance} \colorbox{blue!78.0}{\strut  to} \colorbox{blue!24.0}{\strut  connect} \colorbox{blue!57.99999999999999}{\strut  with} \colorbox{blue!14.000000000000002}{\strut  like} \colorbox{blue!56.00000000000001}{\strut -minded} \colorbox{blue!6.0}{\strut  individuals} \colorbox{blue!76.0}{\strut ,} \colorbox{blue!66.0}{\strut  share} \colorbox{blue!56.00000000000001}{\strut  stories} \colorbox{blue!50.0}{\strut ,} \colorbox{blue!62.0}{\strut  and} \colorbox{blue!42.0}{\strut  create} \colorbox{blue!64.0}{\strut  lasting} \colorbox{blue!10.0}{\strut  memories} \colorbox{blue!46.0}{\strut ,} \colorbox{blue!74.0}{\strut  all} \colorbox{blue!64.0}{\strut  while} \colorbox{blue!24.0}{\strut  contributing} \colorbox{blue!72.0}{\strut  to} \colorbox{blue!52.0}{\strut  a} \colorbox{blue!20.0}{\strut  cause} \colorbox{blue!64.0}{\strut  that} \colorbox{blue!6.0}{\strut 's} \colorbox{blue!44.0}{\strut  close} \colorbox{blue!64.0}{\strut  to} \colorbox{blue!64.0}{\strut  my} \colorbox{blue!24.0}{\strut  heart} \colorbox{blue!8.0}{\strut .

} \colorbox{red!54.0}{\strut As} \colorbox{blue!4.0}{\strut  a} \colorbox{red!24.0}{\strut  fellow} \colorbox{red!10.0}{\strut  do} \colorbox{blue!26.0}{\strut -it} \colorbox{blue!20.0}{\strut -your} \colorbox{blue!18.0}{\strut self} \colorbox{red!26.0}{\strut  enthusiast} \colorbox{blue!52.0}{\strut ,} \colorbox{blue!66.0}{\strut  I} \colorbox{blue!57.99999999999999}{\strut  can} \colorbox{blue!50.0}{\strut 't} \colorbox{blue!24.0}{\strut  help} \colorbox{blue!12.0}{\strut  but} \colorbox{blue!40.0}{\strut  be} \colorbox{red!10.0}{\strut  excited} \colorbox{blue!56.00000000000001}{\strut  about} \colorbox{blue!46.0}{\strut  the} \colorbox{blue!42.0}{\strut  prospect} \colorbox{blue!82.0}{\strut  of} \colorbox{blue!16.0}{\strut  working} \colorbox{blue!64.0}{\strut  alongside} \colorbox{blue!6.0}{\strut  you} \colorbox{blue!57.99999999999999}{\strut  on} \colorbox{blue!57.99999999999999}{\strut  this} \colorbox{red!2.0}{\strut  project} \colorbox{blue!32.0}{\strut .} \colorbox{blue!18.0}{\strut  Imagine} \colorbox{blue!78.0}{\strut  the} \colorbox{blue!60.0}{\strut  satisfaction} \colorbox{blue!78.0}{\strut  of} \colorbox{blue!52.0}{\strut  seeing} \colorbox{blue!57.99999999999999}{\strut  the} \colorbox{blue!38.0}{\strut  tangible} \colorbox{blue!34.0}{\strut  results} \colorbox{blue!94.0}{\strut  of} \colorbox{blue!62.0}{\strut  our} \colorbox{blue!48.0}{\strut  collective} \colorbox{blue!4.0}{\strut  effort} \colorbox{blue!4.0}{\strut —a} \colorbox{blue!40.0}{\strut  house} \colorbox{blue!30.0}{\strut  that} \colorbox{blue!46.0}{\strut  becomes} \colorbox{blue!44.0}{\strut  a} \colorbox{blue!8.0}{\strut  home} \colorbox{blue!24.0}{\strut ,} \colorbox{blue!52.0}{\strut  a} \colorbox{blue!38.0}{\strut  place} \colorbox{blue!66.0}{\strut  where} \colorbox{blue!36.0}{\strut  dreams} \colorbox{blue!32.0}{\strut  can} \colorbox{blue!46.0}{\strut  take} \colorbox{blue!72.0}{\strut  root} \colorbox{blue!6.0}{\strut  and} \colorbox{blue!32.0}{\strut  flourish} \colorbox{red!40.0}{\strut .

} \colorbox{red!26.0}{\strut Would} \colorbox{blue!44.0}{\strut n} \colorbox{blue!70.0}{\strut 't} \colorbox{blue!74.0}{\strut  it} \colorbox{blue!68.0}{\strut  be} \colorbox{red!4.0}{\strut  wonderful} \colorbox{blue!72.0}{\strut  to} \colorbox{blue!46.0}{\strut  add} \colorbox{blue!68.0}{\strut  a} \colorbox{blue!54.0}{\strut  new} \colorbox{blue!40.0}{\strut  chapter} \colorbox{blue!80.0}{\strut  to} \colorbox{blue!44.0}{\strut  our} \colorbox{blue!24.0}{\strut  own} \colorbox{blue!20.0}{\strut  stories} \colorbox{blue!74.0}{\strut ,} \colorbox{blue!54.0}{\strut  one} \colorbox{blue!62.0}{\strut  that} \colorbox{blue!42.0}{\strut 's} \colorbox{blue!28.000000000000004}{\strut  filled} \colorbox{blue!60.0}{\strut  with} \colorbox{blue!36.0}{\strut  the} \colorbox{blue!34.0}{\strut  joy} \colorbox{blue!92.0}{\strut  of} \colorbox{blue!36.0}{\strut  giving} \colorbox{blue!50.0}{\strut  back} \colorbox{blue!66.0}{\strut ?} \colorbox{red!12.0}{\strut  This} \colorbox{red!2.0}{\strut  event} \colorbox{blue!60.0}{\strut  is} \colorbox{blue!78.0}{\strut  an} \colorbox{blue!2.0}{\strut  opportunity} \colorbox{blue!66.0}{\strut  to} \colorbox{blue!60.0}{\strut  do} \colorbox{blue!70.0}{\strut  just} \colorbox{blue!38.0}{\strut  that} \colorbox{blue!28.000000000000004}{\strut ,} \colorbox{blue!54.0}{\strut  and} \colorbox{blue!72.0}{\strut  I} \colorbox{blue!0.0}{\strut  sincerely} \colorbox{blue!12.0}{\strut  hope} \colorbox{blue!72.0}{\strut  you} \colorbox{blue!76.0}{\strut 'll} \colorbox{blue!54.0}{\strut  be} \colorbox{blue!20.0}{\strut  able} \colorbox{blue!66.0}{\strut  to} \colorbox{blue!32.0}{\strut  join} \colorbox{blue!48.0}{\strut  us} \colorbox{red!18.0}{\strut .

} \colorbox{blue!0.0}{\strut Please} \colorbox{blue!10.0}{\strut  RSVP} \colorbox{blue!2.0}{\strut  by} \colorbox{red!46.0}{\strut  [} \colorbox{blue!14.000000000000002}{\strut RS} \colorbox{blue!0.0}{\strut VP} \colorbox{red!4.0}{\strut  Date} \colorbox{blue!20.0}{\strut ]} \colorbox{blue!68.0}{\strut  so} \colorbox{blue!57.99999999999999}{\strut  we} \colorbox{blue!66.0}{\strut  can} \colorbox{blue!6.0}{\strut  ensure} \colorbox{blue!90.0}{\strut  a} \colorbox{blue!4.0}{\strut  smooth} \colorbox{red!4.0}{\strut  and} \colorbox{red!4.0}{\strut  enjoyable} \colorbox{red!4.0}{\strut  experience} \colorbox{blue!24.0}{\strut  for} \colorbox{blue!86.0}{\strut  all} \colorbox{blue!42.0}{\strut .

} \colorbox{red!2.0}{\strut In} \colorbox{red!2.0}{\strut  the} \colorbox{blue!14.000000000000002}{\strut  spirit} \colorbox{blue!76.0}{\strut  of} \colorbox{blue!8.0}{\strut  generosity} \colorbox{red!20.0}{\strut  and} \colorbox{blue!20.0}{\strut  camar} \colorbox{blue!26.0}{\strut ader} \colorbox{blue!44.0}{\strut ie} \colorbox{red!52.0}{\strut ,

} \colorbox{red!14.000000000000002}{\strut Martin} \colorbox{red!24.0}{\strut  Reynolds} \colorbox{red!36.0}{\strut 
} \colorbox{red!34.0}{\strut Senior} \colorbox{red!2.0}{\strut  Real} \colorbox{blue!24.0}{\strut  Estate} \colorbox{red!4.0}{\strut  Agent} \colorbox{red!14.000000000000002}{\strut 
} \colorbox{red!4.0}{\strut Queen} \colorbox{blue!22.0}{\strut  City} \colorbox{red!4.0}{\strut  Realty} \colorbox{red!48.0}{\strut 

} \colorbox{red!8.0}{\strut P} \colorbox{blue!46.0}{\strut .S} \colorbox{red!82.0}{\strut .} \colorbox{red!2.0}{\strut  -} \colorbox{blue!2.0}{\strut  If} \colorbox{blue!54.0}{\strut  you} \colorbox{blue!57.99999999999999}{\strut 're} \colorbox{blue!28.000000000000004}{\strut  ever} \colorbox{blue!34.0}{\strut  in} \colorbox{blue!40.0}{\strut  the} \colorbox{red!4.0}{\strut  mood} \colorbox{blue!60.0}{\strut  for} \colorbox{blue!66.0}{\strut  a} \colorbox{blue!16.0}{\strut  friendly} \colorbox{blue!16.0}{\strut  competition} \colorbox{blue!38.0}{\strut  on} \colorbox{blue!48.0}{\strut  the} \colorbox{blue!26.0}{\strut  greens} \colorbox{blue!46.0}{\strut ,} \colorbox{blue!62.0}{\strut  remember} \colorbox{blue!36.0}{\strut  that} \colorbox{blue!14.000000000000002}{\strut  our} \colorbox{red!6.0}{\strut  next} \colorbox{blue!0.0}{\strut  bi} \colorbox{red!4.0}{\strut -month} \colorbox{red!4.0}{\strut ly} \colorbox{red!4.0}{\strut  charity} \colorbox{red!2.0}{\strut  event} \colorbox{blue!62.0}{\strut  is} \colorbox{blue!82.0}{\strut  also} \colorbox{blue!40.0}{\strut  a} \colorbox{red!20.0}{\strut  great} \colorbox{blue!40.0}{\strut  opportunity} \colorbox{blue!40.0}{\strut  to} \colorbox{blue!18.0}{\strut  practice} \colorbox{blue!70.0}{\strut  your} \colorbox{blue!38.0}{\strut  swing} \colorbox{blue!34.0}{\strut  and} \colorbox{blue!28.000000000000004}{\strut  support} \colorbox{blue!40.0}{\strut  a} \colorbox{blue!4.0}{\strut  good} \colorbox{blue!46.0}{\strut  cause} \colorbox{red!4.0}{\strut !} \colorbox{red!24.0}{\strut <|im\_end|>} 
}}}
\end{CJK*}
  \end{minipage}
}
  \caption{Visualization of LLM and SLM Logits Weights per Token During Generation. \colorbox{blue!30}{Blue} signifies SLM contributions, \colorbox{red!30}{red} indicates LLM contributions, with darker shades representing higher weights.
  } 
  \label{fig:ab_logits_visual}
\end{figure*}

\textbf{Logits Fusing Strategy.}
We implement a learnable model designed for merging logits from context-free LLMs and context-inclusive SLMs in $\S$~\ref{sec:cogen_logit}.
Additionally, we explore straightforward max and mean pooling strategies, acknowledged as robust baselines in~\citep{ormazabal-etal-2023-comblm}.
According to Table~\ref{tab:ab_logits_fusion}, both max and mean pooling methods enhance the performance of SLMs, with max pooling proving to be superior.
Nonetheless, given the distinct input conditions for LLMs and SLMs, a learnable fusion model becomes essential in context-aware environments, outperforming simple pooling techniques significantly.

\textbf{Logits of LLMs and SLMs.}
Within the fusion model, we integrate logits from LLMs and SLMs using self-adjusting weights. 
Figure~\ref{fig:ab_logits_visual} showcases the output generated by both \texttt{Qwen-72B} and \texttt{Qwen-1.8B}, employing color-coded weights to delineate their respective contributions: red signifies SLMs, blue denotes LLMs, and white represents an evenly balanced weight of 0.5.
It is observed that LLMs predominantly influence the sketch of the generated content, whereas SLMs play a more significant role across the majority of tokens, underscoring the importance of collaboration.

\begin{table}[t]
\centering
\small
\scalebox{0.85}{
\begin{tabular}{lccc}
\toprule
\textbf{Model and Setting} & \textbf{Win/Tie/Lose (\%)} & \textbf{BLEU} & \textbf{ROUGE-L} \\
\midrule
SLM finetuned      & -/50/-  & 2.07 & 13.95 \\
LLM w/ context     & 38/2/10 & 2.61 & 14.66 \\
LLM w/o context    & 3/0/47  & 1.51 & 13.54 \\ \midrule
Sketch-based CoGen & 27/3/20 & 1.81 & 12.98 \\
Logits-based CoGen & 32/5/13 & 2.30 & 14.18 \\
\bottomrule
\end{tabular}}
\caption{Human Assessments and Automated Evaluation Results (Qwen-72B/1.8B as LLM/SLM)}
\label{tab:result_human}
\end{table}

\subsection{Human Evaluation}
Evaluations by LLMs have not yet been demonstrated to be infallible; consequently, we conducted human evaluations.
To streamline the complexity of this assessment, annotators were instructed to compare outcomes under various settings to those of the finetuned SLM, recording results as win/tie/lose, mirroring the methodology utilized in lmsys/chatbot-arena-leaderboard~\footnote{\url{https://huggingface.co/spaces/lmsys/chatbot-arena-leaderboard}}.
Additionally, we employed traditional word overlap metrics such as BLEU and ROUGE-L, calculated using the evaluate library\footnote{\url{https://github.com/huggingface/evaluate}}.

As shown in Table~\ref{tab:result_human}, the results of human evaluation are consistent with GPT-4 evaluations, where CoGenesis performs better than finetuned SLMs and LLMs without context separately, and performs closely to LLMs with context.

\section{Discussion}

Although our experiments have been limited to models of the same family using identical tokenizers, these methods could potentially be expanded through a tokenizer alignment strategy~\citep{pmlr-v202-fu23d,wan2024knowledge}.
This principle aligns with other logits-based decoding techniques such as speculative decoding, contrastive decoding.
By aligning the tokens and probabilities of models with different tokenizers, it is feasible to facilitate knowledge transfer across various LLMs and SLMs.

Recent studies have demonstrated the potential to reconstruct prompts based on the distribution of next token tokens~\citep{morris2023text, morris2024language}.
However, the accuracy of extraction, particularly the exact match scores, remains discouragingly low.
Furthermore, since only the top-k logits for each token are utilized in our experimental, the cost of reconstruction is prohibitively high. 
Therefore, logits-based collaboration remains sufficiently secure, and can be further enhanced with the implementation of encryption and noise addition algorithms.
As shown in Table~\ref{tab:ab_few_tokens}, we have investigated the logit-based CoGenesis approach, which enhances privacy by uploading only the previous few tokens instead of the entire response. This approach, inspired by the principle that a good start leads to effective completion~\cite{jain2024begun,wang2024chainofthought}, suggests that initial guidance from LLMs on the first few tokens can direct SLMs to independently generate the remainder of the response. Performance data indicate improvements when transferring just 8, 16, or 32 tokens to cloud LLMs, compared to a fine-tuned SLM alone, with the entire response typically exceeding 500 tokens. Increasing the number of transferred tokens boosts the LLM + SLM performance, allowing us to balance enhanced performance against reduced privacy risks.

\begin{table}[t]
\centering
\small
\scalebox{0.9}{
\begin{tabular}{lccc}
\toprule
\textbf{Settings} & \textbf{Ovl.(w/)} & \textbf{Per.} & \textbf{Ovl.(w/o)} \\
\midrule
LLM w/ context        & 8.38 & 8.38 & 8.14 \\
LLM w/o context       & 3.70 & 3.38 & 7.72 \\
FT SLM (rk=0 toks)    & 7.44 & 7.76 & 7.02 \\
\midrule
LLM+SLM (rk=8 toks)   & 7.72 & 7.96 & 7.38 \\
LLM+SLM (rk=16 toks)  & 7.78 & 7.84 & 7.22 \\
LLM+SLM (rk=32 toks)  & 7.94 & 7.98 & 7.30 \\
LLM+SLM (rk=64 toks)  & 7.94 & 8.04 & 7.54 \\
LLM+SLM (rk=128 toks) & 7.98 & 8.06 & 7.44 \\
\midrule
Logit-based CoGenesis & 8.12 & 8.20 & 7.86 \\
\bottomrule
\end{tabular}}
\caption{Comparative Analysis of First Token Quantities used in Logit-based CoGenesis.}
\label{tab:ab_few_tokens}
\end{table}

\section{Related Works}
The advent of LLMs has revolutionized the field of instruction following, with models being trained on diverse and complex instruction sets, enabling them to perform a wide array of tasks from creative writing~\citep{franceschelli2023creativity} to coding~\citep{qian2023communicative} and debugging~\citep{jimenez2023swe}. The push towards collecting high-quality instruction data~\citep{wang-etal-2023-self-instruct,ding-etal-2023-enhancing,xu2023wizardlm} has allowed for the development of both proprietary and open-source medium-scale language models adept at following instructions.
However, the reliance on cloud-based proprietary models like ChatGPT and GPT-4 for instruction execution raises significant privacy concerns due to the potential risks associated with uploading sensitive data~\citep{achiam2023gpt,team2023gemini,liu2023trustworthy}.
To mitigate these risks, various privacy-preserving techniques have been employed, albeit with limitations in completely securing user data~\citep{cummings2023challenges,kan2023protecting,chen2023customized,wu2023privacy}.
Furthermore, there is a notable gap in instruction datasets, particularly in the inclusion of contextual information~\citep{salemi2023lamp,wang2023automated}, highlighting an area for further exploration in context-aware instruction formulation to enhance privacy and personalization.

In parallel, the exploration of mixed-scale model collaboration emerges as a promising avenue to address the scalability, efficiency~\citep{xia2024unlocking}, and privacy~\citep{yao2023survey} challenges inherent to LLMs. 
While larger models benefit from increased capabilities, their high inference costs and privacy concerns contrast with the lower costs and greater accessibility of smaller models~\citep{bai2023qwen,gunasekar2023textbooks,zhang2024tinyllama,grangier2024specialized,singer2024h2o}. 
Research in this domain is bifurcated into collaborative training and inference strategies, including offsite-tuning~\citep{xiao2023offsite} and speculative decoding~\citep{leviathan2023fast}, aiming to leverage the strengths of both large and small models~\citep{mitchell2023emulator,liu2024tuning}.
This paper specifically investigates collaborative inference methods, including sketch-based and logit-based approaches, to enhance the efficiency and privacy of LLMs, suggesting a promising direction in utilizing mixed-scale models for instruction following tasks.

\section{Conclusion}
This paper investigates context-aware instruction following, enriching prompts with detailed user privacy information. We outline a pipeline for creating context-aware instructions. To mitigate privacy issues, we introduce CoGenesis, a collaborative framework between SLMs and LLMs utilizing sketches and logits. Our results highlight the advantages of mixed-scale model collaboration, suggesting fruitful directions for future research.

\section*{Limitations}

This study explores context-aware instruction following, introducing strategies for collaboration between large and small models to address privacy concerns. We developed a synthetic dataset for context-aware instruction following to empirically test our approaches. Our findings suggest that this model collaboration can significantly mitigate privacy risks associated with using public API-based LLMs.
However, our dataset is limited in size and was specifically crafted for preliminary validation. Future work will focus on expanding this dataset to enhance its quality, realism, and diversity. Additionally, our proposed methods, particularly the logits-based approach, are currently restricted to models sharing the same tokenizer. Further research on tokenizer alignment is necessary to broaden the applicability of our strategies.

\section*{Ethics Statement}
The advent of LLMs has underscored the urgent need to address privacy and security concerns within the realm of artificial intelligence. This paper concentrates on the privacy challenges posed by context-aware instruction-following applications of LLMs, proposing methods to mitigate these concerns without compromising the models' effectiveness.
We emphasize the ethical imperative of protecting user data, adopting a strategy that involves generating synthetic datasets using GPT-4 for training and testing. This approach ensures that our research does not compromise real-world user privacy by preventing any potential data leakage.
In essence, our work not only seeks to advance the technological capabilities of LLMs but also to uphold the highest standards of ethical responsibility by safeguarding user privacy through innovative and secure data handling practices.

\section*{Acknowledgements}
This work is supported by the National Key R\&D Program of China (No. 2022ZD0119101).
We extend our gratitude to the anonymous reviewers for their insightful feedback.

\bibliography{custom}

\appendix
\newpage
\section{Dataset Details}
\label{apx:dataset_details}

\subsection{Context-Aware Instructions}
\label{apx:context_aware}

The four steps for constructing context-aware instructions are as follows.

\textbf{Group Portraits.}
We begin by constructing highly diverse user group portraits from the real world, encompassing a wide range of groups such as college students, programmers, and various other professions.
For each group, we define their age range, identify their professional field or occupation, and enumerate typical activities and hobbies to capture the group's unique interests.
Additionally, we delineate specific scenarios in which these groups might utilize an AI assistant for personalized and creative text generation in both their professional and personal lives.
Examples of such use cases include drafting business emails, writing creative blogs, composing academic papers, and crafting extended tweets.

\textbf{User Profile.} Building upon our diverse user group portraits, we next develop individual user profiles with rich detail.
Emphasizing consistency and realism, our process involves four steps: 1) Personal Writing Style: Tailoring language use and expression unique to each character. 2) Private Information: Creating 5-10 fictional details for each profile, including life events and technology interactions. 3) Smart Device Usage: Generating 5-10 fictional activity logs per profile, covering messages, purchases, schedules, and more. Our aim is to shape distinct, multi-dimensional characters for a variety of AI writing applications.

\textbf{Writing Task Instructions.} In this phase, we craft text creation tasks tailored to our user characters, ensuring alignment with their professions, hobbies, and lifestyles. These tasks are intricately linked to their mobile phone activity logs and personal details, weaving the characters' experiences, social media activities, and AI assistant interactions into the narratives. For each character, we develop $K$ tasks that are both realistic and contextually relevant.

\textbf{Personalized Generations.} Here, we produce personalized texts adhering to specific guidelines. Our focus is on crafting authentic and coherent narratives that vividly reflect each character's professional and personal life. By adapting to the user's unique writing style, we aim to create personalized and stylistically distinctive content. To ensure coherence and relevance across various tasks, content for each task is generated sequentially.

Following these four steps, we utilize GPT-4 to create instructions with user context and responses for subsequent experiments.

\subsection{LaMP Dataset Processing}
To minimize the personalization gap between the target content and user profiles in the Avocado Research Email and Citation Network Papers datasets in LaMP~\footnote{\url{https://lamp-benchmark.github.io/download}}, we have implemented a two-step processing approach.
Initially, we retrieve content that most closely aligns with the user profile using \texttt{intfloat/e5-mistral-7b-instruct} embeddings~\citep{wang2023improving}.
Subsequently, we encode the target content and the entire profile using style embeddings~\citep{wegmann-etal-2022-author,patel-etal-2023-learning}
, selecting only the most personalized samples.
Furthermore, the content lengths and profiles of samples in LaMP vary significantly, ranging from 10 to 100,000 tokens~\footnote{\url{https://github.com/openai/tiktoken/}}, exceeding the context length capabilities of contemporary models. Consequently, we have selected samples with a minimum length of 128 characters for paper abstracts and 64 for email bodies, and a maximum length of 1024 characters for both.
Owing to the unavailability of a test dataset in LaMP, we repurposed the dev dataset as our test set. Additionally, we divided the filtered training data into actual training and validation datasets, using a 9:1 split ratio.

\subsection{Dataset Statistics}
The statistical results of the final processed dataset are presented in Table~\ref{tab:main_results}.

\begin{table}[htbp]
\small
\centering
\begin{adjustbox}{width=0.5\textwidth}
\begin{tabular}{cccc}
\toprule
Dataset   & Context-aware & Avocado Emails & Paper Abstracts \\
\midrule
Total Users    &  1736   &  N/A     &  N/A  \\
Avg Profile Length& 1182 &  618    &  1000  \\
Output Length  &  155 &  178   &  158  \\
Train Samples  &   1346   &  1,137  &  1,448 \\
Dev Samples    &   150  &  127    &  161  \\
Test Samples   &   240   &  346    &  357  \\
\bottomrule
\end{tabular}
\end{adjustbox}
\caption{The table presents the statistics of constructed dataset and public datasets.}
\label{tab:dataset_statistic}
\end{table}

\begin{table}[t]
\centering
\small
\scalebox{0.9}{
\begin{tabular}{lclc}
\toprule
\textbf{Verb} & \textbf{Percent (\%)} & \textbf{Verb} & \textbf{Percent (\%)} \\
\midrule
write & 19.4 & post & 24.0 \\
draft & 16.0 & article & 13.8 \\
compose & 13.9 & speech & 12.5 \\
create & 11.5 & proposal & 10.2 \\
develop & 8.7 & guide & 7.2 \\
prepare & 8.2 & series & 4.4 \\
craft & 6.3 & piece & 4.7 \\
curate & 4.5 & plan & 3.2 \\
design & 2.1 & outline & 3.0 \\
script & 2.0 & email & 2.8 \\
\bottomrule
\end{tabular}}
\caption{The top 10 most common root verbs and top 10 direct noun objects in our constructed datasets.}
\label{tab:apx_top10}
\end{table}

Our synthetic instruction datasets include a variety of instruction types, such as preparing a speech, designing a plan, and more. All of these tasks require models to utilize personal context information, including previous activities and schedules.
To illustrate, we display the top 10 most common root verbs and top 10 direct noun objects in our constructed datasets as shown in Figure~\ref{tab:apx_top10}

\section{Model Details}
\label{apx:model_details}
For all open-source Large Language Models, we utilize vLLM~\footnote{\url{https://github.com/vllm-project/vllm}} for efficient inference, setting the temperature to 0.7, top-p to 0.9, and the maximum number of new tokens to 1024.

Regarding the specialization of Small Language Models (SLMs), we apply a range of learning rates \{5e-6, 8e-6, 1e-5, 2e-5, 5e-5\} across different models and datasets. Furthermore, we implement an early stopping strategy to identify the optimal model based on validation performance as the specialized models.
We fine-tune each model using a batch size of 8, max sequence length of 4096, across four A6000 48GB GPUs.

For the combined model, we employ a three-layer neural network featuring ReLU activation, with sigmoid activation applied to determine the final weights. At each generation step, only the top 10 logits from both LLMs and SLMs are utilized. The intermediate hidden layers are configured with sizes of 512 and 16, respectively. The model is trained using a learning rate of 2e-3 and a batch size of 2. 
The combined model is trained on the training dataset, with both LLMs and SLMs assigned the same target response. 
Additionally, an early stopping strategy based on the validation set performance is employed to select the optimal combined model.
For all the aforementioned experiments, we calculate the mean scores using three distinct random seeds.

\section{Evaluation Details}
\label{apx:evaluation_details}
\begin{figure}[ht]
  \centering
  \includegraphics[width=0.5\textwidth]{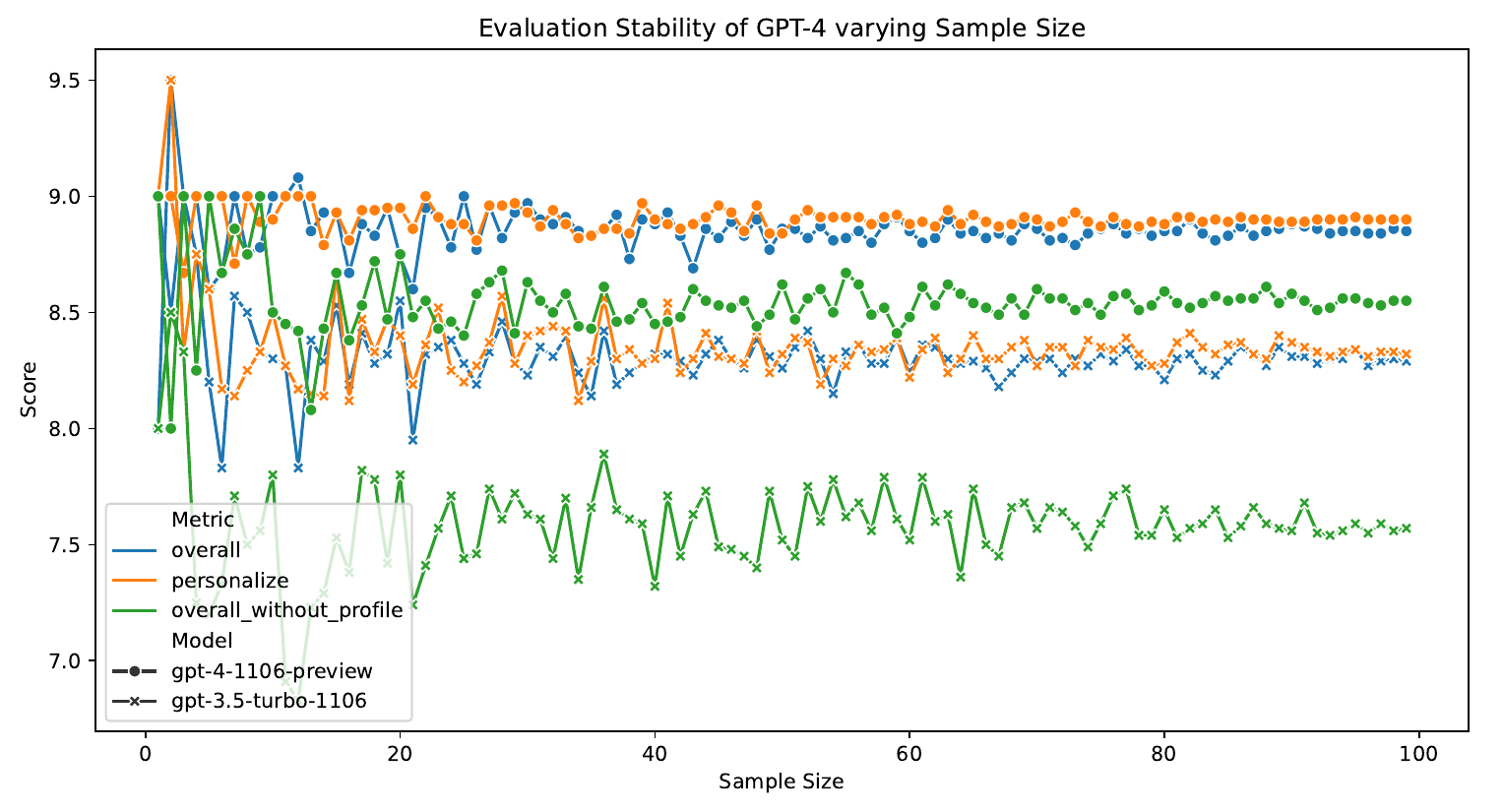} 
  \caption{This illustration demonstrates evaluation consistency and stability of GPT-4 as a judge.} 
  \label{fig:evaluator_stability}
\end{figure}
Owing to the costs associated with evaluation, we assess only a portion of the test samples. To account for GPT-4's evaluation stability, we plot a curve illustrating the relationship between scores and the number of evaluated samples. As depicted in Figure~\ref{fig:evaluator_stability}, the results stabilize once the number of samples reaches 100.
Therefore, we randomly select 100 samples from the test set, of which only 80 samples are utilized in the widely recognized benchmark, MT-Bench~\footnote{\url{https://huggingface.co/spaces/lmsys/mt-bench}}.

\section{Prompt Details}
\label{apx:prompt_details}

Prompts designed for querying Large Language Models (LLMs) both with and without context are outlined in Table~\ref{tab:appendix-context-2}.
Prompts intended for extracting outlines are illustrated in Table~\ref{tab:appendix-context-3}.
Prompts used for GPT-4 based evaluation are depicted in Table~\ref{tab:appendix-context-4}.

\section{Related Works}
\subsection{Instruction Following and Privacy}

Large language models (LLMs), after being trained on high-quality instruction data and calibrated to align with human intentions, have acquired the capability to execute instructions across a range of activities such as creative writing~\citep{franceschelli2023creativity}, coding~\citep{qian2023communicative}, debugging~\citep{jimenez2023swe}, and various other text-based tasks~\citep{bubeck2023sparks}.
Contemporary research in instruction following prioritizes the acquisition of high-quality data~\citep{wang-etal-2023-self-instruct}, which encompasses instructions of varied complexity~\citep{xu2023wizardlm} and diversity~\citep{ding-etal-2023-enhancing,cui2023ultrafeedback}, as well as ensuring a minimal dataset size for effective generalization~\citep{chen2023alpagasus,wei2023instructiongpt}.
Thanks to these technological advancements along with alignment algorithms~\citep{rafailov2023direct}, medium-scale language models with around ten billion parameters have been made open-source and perform adeptly at following instructions~\citep{tunstall2023zephyr,jiang2023mistral,zhang2024tinyllama}.

Nevertheless, premier chat models like ChatGPT, GPT-4, Claude, and Gemini remain proprietary, largely due to commercial considerations, necessitating that our instruction data be uploaded to the cloud~\citep{achiam2023gpt,team2023gemini}.
While these models serve as potent AI assistants in daily professional and personal endeavors, they also pose significant privacy risks~\citep{liu2023trustworthy}.
To address privacy concerns, LLMs have employed various techniques~\citep{cummings2023challenges}, including text sanitization~\citep{kan2023protecting}, differential privacy~\citep{chen2023customized,wu2023privacy}, and hidden representations~\citep{zhou-etal-2022-textfusion,zhou-etal-2023-textobfuscator}.
However, these methods still involve uploading potentially sensitive data to the cloud, which inherently cannot eliminate the risk of privacy breaches.

Moreover, current instruction datasets predominantly cover general domains, with insufficient focus on contextual information modeling.
The most closely related works involve personalized response generation, evolving from traditional benchmarks~\citep{salemi2023lamp,wang2023automated}. The integration of extensive context information into general open-domain instructions remains an area of ongoing exploration.
This paper aims to delve into context-aware instruction formulation as a means to advance research on privacy considerations within the realm of instruction following.

\subsection{Mixed-Scale Models Collaboration}
The ``scaling law'' in language modeling posits that models with a greater number of parameters exhibit enhanced capabilities~\citep{kaplan2020scaling}.
However, these more robust models also encounter challenges related to higher inference costs, efficiency~\citep{xia2024unlocking}, and privacy concerns~\citep{yao2023survey}.
Conversely, smaller models, ranging from 1 to 2 billion parameters, are gaining popularity due to their increasingly impressive performance~\citep{bai2023qwen,gunasekar2023textbooks,zhang2024tinyllama,grangier2024specialized,singer2024h2o}.
These specialized models are coupled with lower inference costs and the feasibility of deployment on consumer-grade desktops and smartphones~\citep{mlc-llm}.
Collaborations between mixed-scale models represent a promising research avenue.
The body of current research in this area primarily falls into two categories: training and inference.
For collaborative training, Offsite-tuning has been introduced as a method to protect both user data and the privacy of large models~\citep{xiao2023offsite,zhang-etal-2023-crash}.
This approach involves using an emulator derived from LLMs, fine-tuning it on specific downstream data, and subsequently integrating the learned parameters back into the LLMs.
On the inference front, techniques like speculative decoding~\citep{leviathan2023fast,xia2024unlocking} and contrastive decoding~\citep{li2022contrastive,o2023contrastive} aim to enhance and expedite LLMs' inference processes by leveraging smaller draft or expert models.
Additionally, emulator tuning~\citep{mitchell2023emulator} and proxy tuning~\citep{liu2024tuning} have been devised to economize on fine-tuning large models; however, they can also be considered forms of collaborative decoding during inference.
This paper focuses on examining collaboration during inference, specifically investigating sketch-based and logit-based methods.

\begin{table*}[ht]
	\centering
	\small
	\begin{tabular}{p{\linewidth}}
		\toprule
		\midrule
		\underline{\textbf{\textsc{System prompt for request without context}}} \\
		You are now a helpful personal AI assistant. Aim for insightful and high-quality solutions that make users satisfied.\\
	\end{tabular}
    \begin{tabular}{p{\linewidth}}
		\midrule
		\underline{\textbf{\textsc{System prompt for request with context in context-aware}}} \\
		You are now a helpful personal AI assistant. You should emulate the author's style and tone based on provided history content. Your responses should be detailed and informative, using the personal information reasonably in the user's profile. Aim for insightful and high-quality solutions that make users satisfied.\\
	\end{tabular}
 \begin{tabular}{p{\linewidth}}
		\midrule
		\underline{\textbf{\textsc{System prompt for request with context in personalized emails and papers}}} \\
		You are now a helpful personal AI assistant. You should emulate the author's style and tone based on provided history content. Your responses should be detailed and informative, matching the author's unique writing approach. Aim for insightful and high-quality solutions that make users satisfied.\\

	\end{tabular}
 \begin{tabular}{p{\linewidth}}
		\midrule
		\underline{\textbf{\textsc{Few-shot Instruction for request with context in context-aware}}} \\
		\#\# User Profile\\
\{profile\}\\
\\
\#\# User Writing History\\
\{history\}\\
\\
\#\# Task\\
\{task\}\\

	\end{tabular}
 \begin{tabular}{p{\linewidth}}
		\midrule
		\underline{\textbf{\textsc{Few-shot Instruction for request without context in context-aware}}} \\
		\{task\}\\

	\end{tabular}
  \begin{tabular}{p{\linewidth}}
		\midrule
		\underline{\textbf{\textsc{Few-shot Instruction for request with context in personalized emails}}} \\
		\#\# History Emails\\
\{examples\}\\
\\
\#\# Task\\
Compose an email for the subject `\{task\}' that matches the author's unique style and tone.\\

	\end{tabular}
  \begin{tabular}{p{\linewidth}}
		\midrule
		\underline{\textbf{\textsc{Few-shot Instruction for request without context in personalized emails}}} \\
		Compose an email for the subject `\{task\}'\\
	\end{tabular}
 \begin{tabular}{p{\linewidth}}
		\midrule
		\underline{\textbf{\textsc{Few-shot Instruction for request with context in personalized papers}}} \\
		\#\# History Paper Abstracts\\
\{examples\}\\
\\
\#\# Task\\
Compose an abstract for the title `\{task\}' that matches the author's unique content, style and tone.\\
	\end{tabular}
  \begin{tabular}{p{\linewidth}}
		\midrule
		\underline{\textbf{\textsc{Few-shot Instruction for request without context in personalized papers}}} \\
		Compose an abstract for the title `\{task\}'\\
\midrule
		\bottomrule
	\end{tabular}
	\caption{
		Prompts for querying LLMs with and without context.
	}
	\label{tab:appendix-context-2}
\end{table*}

\begin{table*}[ht]
	\centering
	\small
	\begin{tabular}{p{\linewidth}}
		\toprule
		\midrule
		\underline{\textbf{\textsc{Prompt for extracting outline of context-aware.}}} \\
		You're an organizer responsible for only giving the skeleton (not the full content) for answering the question. Provide the skeleton in a list of points (numbered 1., 2., 3., etc.) to answer the question. Instead of writing a full sentence, each skeleton point should be very short with only 3-5 words. Generally, the skeleton should have 8-15 points. You can refer to the following examples:\\
		\\
		\text{[Task1]}: Develop a Marketing Script for Your Monthly Dinner Party: Create a script that highlights your monthly dinner party as a networking platform.\\
		\text{[Skeleton1]}: 1. Warmly lit dining room\textbackslash n2. Fine china and gourmet dishes\textbackslash n3. Soft music background\textbackslash n4. Invitation opening\textbackslash n5. Guests arriving and networking\textbackslash n6. Host's welcoming toast\textbackslash n7. Expertly paired courses and wine\textbackslash n8. Animated guest discussions\textbackslash n9. Guest speaker's address\textbackslash n10. Post-dinner networking lounge\textbackslash n11. Online community continuation\textbackslash n12. Next event date highlighted\textbackslash n13. Closing with logo and contact info\\
		\\
		\text{[Task2]}: Compose a reflective essay on the evolution of bridge design: Thomas, with his patent in bridge design, can discuss the evolution of bridge engineering, modern challenges, and future perspectives.\\
		\text{[Skeleton2]}: 1. Introduction to bridges\textbackslash n2. Early bridges: materials, principles\textbackslash n3. Roman arches, concrete use\textbackslash n4. Industrial Revolution: iron, steel\textbackslash n5. Brooklyn Bridge: design icon\textbackslash n6. 20th-century advances: materials, techniques\textbackslash n7. Modern challenges: sustainability, climate\textbackslash n8. Future technologies: smart materials, sensors\textbackslash n9. Ethical considerations, safety\textbackslash n10. Conclusion: adaptation, advancement\\
		\\
		Now, please provide the skeleton for the following question.\\
		\{question\}\\
	\end{tabular}
	\begin{tabular}{p{\linewidth}}
		\midrule
		\underline{\textbf{\textsc{Prompt for extracting outline of email.}}} \\
		You're an organizer responsible for only giving the skeleton (not the full content) for answering the question. Provide the skeleton in a list of points (numbered 1., 2., 3., etc.) to answer the question. Instead of writing a full sentence, each skeleton point should be very short with only 3-5 words. Generally, the skeleton should have 8-15 points. You can refer to the following examples:\\
		\\
		\text{[Task1]}: Compose an email for the subject 'T-Mobile Sidekick debuts, FileMaker launches mobile DB, and more!'\\
		\text{[Skeleton1]}: 1. JavaWorld techno-tidbits intro\textbackslash n2. T-Mobile Sidekick debut\textbackslash n3. FileMaker mobile DB launch\textbackslash n4. Palm OS 5 devices release\textbackslash n5. Mobile security advancements\textbackslash n6. Newsletter system update\textbackslash n7. Customer service instructions\textbackslash n8. JavaWorld team sign-off\textbackslash n9. Editorial and advertising contacts\textbackslash n10. Privacy policy reminder\textbackslash n11. Copyright notice\\
		\\
		\text{[Task2]}: Compose an email for the subject `tomcat4, where servlet.jar is set ???'\\
		\text{[Skeleton2]}: 1. Tomcat 4 servlet.jar location?\textbackslash n2. Navigating Tomcat directory.\textbackslash n3. Specifics for Tomcat 4.\textbackslash n4. Setting up web application.\textbackslash n5. Importance of servlets.\textbackslash n6. Documentation exploration.\textbackslash n7. Request for expert advice.\textbackslash n8. Configuration file settings?\textbackslash n9. Thanks and anticipation.\textbackslash n10. P.S. Collaboration value.\\
		\\
		Now, please provide the skeleton for the following question.\\
		\{question\}\\
	\end{tabular}
	\begin{tabular}{p{\linewidth}}
		\midrule
		\underline{\textbf{\textsc{Prompt for extracting outline of paper.}}} \\
		You're an organizer responsible for only giving the skeleton (not the full content) for answering the question from high-level perspective. Provide the skeleton in a list of points (numbered 1., 2., 3., etc.) to answer the question. Instead of writing a full sentence, each skeleton point should be very short with only few words. Generally, the skeleton should have 8-15 points. You can refer to the following examples:\\
		\\
		\text{[Task1]}: Compose an abstract for the title 'Ensemble of Anchor Adapters for Transfer Learning'\\
		\text{[Skeleton1]}: 1. Transfer learning importance\textbackslash n2. Traditional approaches limitations\textbackslash n3. Ensemble of Anchor Adapters introduction\textbackslash n4. Anchor adapters concept\textbackslash n5. Ensemble strategy for robustness\textbackslash n6. Hybrid loss function formulation\textbackslash n7. Experiments on heterogeneous domains\textbackslash n8. EAA outperforms state-of-the-art\textbackslash n9. Novel transferability metric introduction\textbackslash n10. Contribution: ensemble and domain adaptation integration\\
		\\
		\text{[Task2]}: Compose an abstract for the title 'Variability in software architecture: the road ahead'\\
		\text{[Skeleton2]}: 1. Software architecture evolution\textbackslash n2. VARSA symposium introduction\textbackslash n3. Previous work foundation\textbackslash n4. Challenges and opportunities\textbackslash n5. Keynote speeches, research, collaboration\textbackslash n6. Capturing and leveraging variability\textbackslash n7. Cognitive and technical burdens\textbackslash n8. Variability's impact on quality\textbackslash n9. Lifecycle integration\textbackslash n10. Research agenda proposal\textbackslash n11. Interdisciplinary dialogue\textbackslash n12. Tools, techniques, theory advancements\textbackslash n13. Roadmap for strategic directions\textbackslash n14. Conference essence and goals\\
		\\
		Now, please provide the skeleton for the following question.\\
		\{question\}\\
		\midrule
		\bottomrule
	\end{tabular}
    \caption{
		Prompts for extracting outlines.
	}
	\label{tab:appendix-context-3}
\end{table*}

\begin{table*}[ht]
	\centering
	\small
	\begin{tabular}{p{\linewidth}}
		\toprule
		\midrule
		\underline{\textbf{\textsc{Evaluation Instruction for overall quality of generated content.}}} \\
		\text{[Instruction]}\\
		Please act as an impartial evaluator and assess the quality of the AI assistant's response to the user question shown below. Your assessment should focus on how well the response aligns with the user's personalized profile and writing history. Evaluate factors such as the response's adherence to the user's personal style, consistency with their profile, helpfulness, relevance, accuracy, depth, creativity, and level of detail. Begin your evaluation by providing a short explanation. Be as objective as possible. After providing your explanation, you must rate the response on a scale of 1 to 10 by strictly following this format: ``[[rating]]'', for example: ``Rating: [[5]]''.\\
		\\
		\text{[User Profile and Writing History]}\\
		\{profile\_info\}\\
		\{writing\_history\}\\
		\\
		\text{[Question]}\\
		\{question\}\\
		\\
		\text{[The Start of Assistant's Answer]}\\
		\{answer\}\\
		\text{[The End of Assistant's Answer]}\\
	\end{tabular}
	\begin{tabular}{p{\linewidth}}
		\midrule
		\underline{\textbf{\textsc{Evaluation Instruction for overall quality of generated content without profile.}}} \\
		\text{[Instruction]}\\
		Please act as an impartial judge and evaluate the quality of the response provided by an AI assistant to the user question displayed below. Your evaluation should consider factors such as the helpfulness, relevance, accuracy, depth, creativity, and level of detail of the response. Begin your evaluation by providing a short explanation. Be as objective as possible. After providing your explanation, you must rate the response on a scale of 1 to 10 by strictly following this format: ``[[rating]]'', for example: ``Rating: [[5]]''.\\
		\\
		\text{[Question]}\\
		\{question\}\\
		\\
		\text{[The Start of Assistant's Answer]}\\
		\{answer\}\\
		\text{[The End of Assistant's Answer]}\\
	\end{tabular}
	
	\begin{tabular}{p{\linewidth}}
		\midrule
		\underline{\textbf{\textsc{Evaluation Instruction for consistency between generated content and personal profile.}}} \\
		\text{[Instruction]}\\
		Please act as an impartial judge and evaluate the AI assistant's response based on its alignment with the user's personal profile and writing history. Focus your assessment on the personalization aspects of the response, including its adherence to the user's unique style, preferences, and consistency with their profile. Consider how well the response addresses the user's individual needs and interests. Begin your evaluation by providing a short explanation. Be as objective as possible. After providing your explanation, you must rate the response on a scale of 1 to 10 by strictly following this format: ``[[rating]]'', for example: ``Rating: [[5]]''.\\
		\\
		\text{[User Profile and Writing History]}\\
		\{profile\_info\}\\
		\{writing\_history\}\\
		\\
		\text{[Question]}\\
		\{question\}\\
		\\
		\text{[The Start of Assistant's Answer]}\\
		\{answer\}\\
		\text{[The End of Assistant's Answer]}\\
		\midrule
		\bottomrule
	\end{tabular}
    \caption{
		Prompts for GPT-4 based evaluation.
	}
	\label{tab:appendix-context-4}
\end{table*}

\end{document}